\documentclass[nohyperref]{article}

\usepackage{microtype}
\usepackage{graphicx}
\usepackage{subfigure}
\usepackage{booktabs} % for professional tables
\usepackage{float}

\usepackage{hyperref}

\usepackage[accepted]{icml2022}

\usepackage{amsmath}
\usepackage{amssymb}
\usepackage{mathtools}
\usepackage{amsthm}

\usepackage[capitalize,noabbrev]{cleveref}

\theoremstyle{plain}

\theoremstyle{definition}

\theoremstyle{remark}

\usepackage[textsize=tiny]{todonotes}

\icmltitlerunning{}

\begin{document}

\twocolumn[
\icmltitle{Deephys: Deep Electrophysiology \\
Debugging Neural Networks under Distribution Shifts}




\begin{icmlauthorlist}
\icmlauthor{Anirban Sarkar}{yyy,cbm}
\icmlauthor{Matthew Groth}{yyy,cbm}
\icmlauthor{Ian Mason}{yyy,cbm}
\icmlauthor{Tomotake Sasaki}{comp,cbm}
\icmlauthor{Xavier Boix}{yyy,cbm}\\ \vspace{.5cm}
\icmlauthor{\url{https://deephys.org}}{}
\end{icmlauthorlist}

\icmlaffiliation{yyy}{Department of Brain and Cognitive Sciences, Massachusetts Institute of Technology, Cambridge, USA}
\icmlaffiliation{cbm}{Center for Brains, Minds \& Machines, Cambridge, USA}
\icmlaffiliation{comp}{Artificial Intelligence Laboratory, Fujitsu Limited, Kawasaki, Japan}

\icmlcorrespondingauthor{Anirban Sarkar}{anirbans@mit.edu}
\icmlcorrespondingauthor{Xavier Boix}{xboix@mit.edu}

\icmlkeywords{Machine Learning}

\vskip 0.3in
]



\printAffiliationsAndNotice{}

\begin{abstract}
Deep Neural Networks (DNNs) often fail in out-of-distribution scenarios.
In this paper, we introduce a tool to visualize and understand such failures. We draw inspiration from concepts from neural electrophysiology, which are based on inspecting the internal functioning of a neural networks by analyzing the feature tuning and invariances of individual units. Deep Electrophysiology, in short Deephys, provides insights of the DNN's failures in out-of-distribution scenarios by comparative visualization of the neural activity in in-distribution and out-of-distribution datasets. Deephys provides seamless analyses of individual neurons, individual images, and a set of set of images from a category, and it is capable of revealing failures due to the presence of spurious features and novel features. We substantiate the validity of the qualitative visualizations of Deephys thorough quantitative analyses using convolutional and transformers architectures, in several datasets and distribution shifts (namely, colored MNIST, CIFAR-10 and ImageNet).

% % Matt's revision
% Matt's revision 

% Despite the success of Deep Learning models in classifying visual and linguistic data, they often struggle in out-of-distribution scenarios
% Taking inspiration from techniques of modern neuroscience, here we introduce a tool for inspecting the behavior of model neurons in these scenarios.
% Deep Electrophysiology, in short Deephys, unravels the internal mechanisms of deep neural networks through comparative visualization of neurons for in distribution (InD) and out of distribution (OoD) datasets. Visualizations focus on the change in tuning and the resultant performance drop that neurons undergo in OoD cases. 
% Deephys can inspect a neuron, image, category, or a pair of categories that are being confused by the model. It provides synchronized analyses according to the neuronal responses to InD and OoD input. 
% As a single biological neuron's firing properties in response to different stimuli are rigorously compared in an electrophysiological study, Deephys' comparative analyses can shine a flashlight into the black box of deep learning and generate insights into how or why a model has dysfunction in OoD cases.
% With Deephys, we inspected a broad range of datasets and models, such as colored MNIST, CIFAR-10 and ImageNet. Here we aim to substantiate the validity and usefulness of the proposed tool.

\end{abstract}

\section{Introduction}
\label{Introduction}

Deep Neural Networks (DNNs) do not generalize well to different distributions from the training data distribution.
The internal DNN mechanisms that lead to out-of-distribution (OOD) failures remains largely unknown. Understanding such mechanisms is a key question for accountability, safety and fairness of intelligent systems.
A promising strand of research for understanding the internal DNNs' decision-making mechanisms that yet has to be applied to understand OOD failures, is based on interpreting the neural activity of individual DNN's neurons.
It is well known that neurons are tuned to detect features present in the training data \cite{zeiler2014visualizing, bau2020understanding, elhage2022superposition}. That is, neurons are selective to a feature as the output value of the neuron is only high when the feature is present in the image. Also, such selectivity is invariant to some nuisance factors in the data, and invariance is at the core of generalization. The emergence of selectivity and invariant neural activity is a well-known phenomenon for both biological and artificial neurons and has already been shown to play a key role for OOD generalization \cite{anselmi2016invariance, sinha1996role, ullman2000high}.

Previous works studied individual neurons in great detail \cite{zeiler2014visualizing, bau2020understanding}, but they are mostly targeted to understanding model behaviors with InD. We extend the effort to examine neurons, for InD as well as OOD, through a neural tuning based explainability technique for visualization, developed over previous efforts \cite{koh2020concept, sarkar2022framework}, that can generate global explanations, and is effective for OOD detection \cite{choi2022concept}.

When presented with out-of-distribution (OOD) input, neurons may exhibit atypical behaviours which correspond to the model's degraded performance. Therefore, insights can be gained at the neural level into the causes, and possible solutions, for failures under distribution shift. Following this hypotheses, we resort to interpreting model behaviours through understanding patterns in the neural tuning for different data distributions, and investigating functional disparity of the neurons to gain insights about biases present in the datasets.

While the prior works, conducted with a controlled set of features or dataset biases, establish new phenomena, they are not applicable for natural datasets due to unavailability of apriori bias factors. 
Such features can be visible yet not easily quantifiable, in contrast to simulated datasets, where the images are generated with proper calculation of the biases.
Here we concentrate on realizing these biases through visualizing difference in behavioural response of individual neurons for different data distributions.
Such exploration provides enlightening insights about how tuning of every neuron changes corresponding to the change in biases from InD to OOD.
In this paper, we focus on explaining distribution shifts through proposing multiple logically connected strategies. They provide meaningful interpretations of inconsistent model behaviors through a new visualization technique, that can conceptualize the distinguishing factors of out-of-distribution (OOD) data compared to in-distribution (InD) data.
Such differentiating traits are also captured by our proposed novel quantitative metrics, that corroborate our study and complement our findings about the model.
Borrowing the terminology `\textit{Electrophysiology}' from neuroscience, which is a study of electrical signals in biological cells and tissues, here we propose a toolkit to study neuron activation signals of deep networks under distribution shift. The idea to look at the response patterns of a single neuron as closely as possible inspired the name `\textit{Deep Electrophysiology}', in short `\textit{Deephys}'.

\begin{figure*}[h]
    \centering
    \begin{tabular}{c}
    \includegraphics[width=1.0\textwidth]{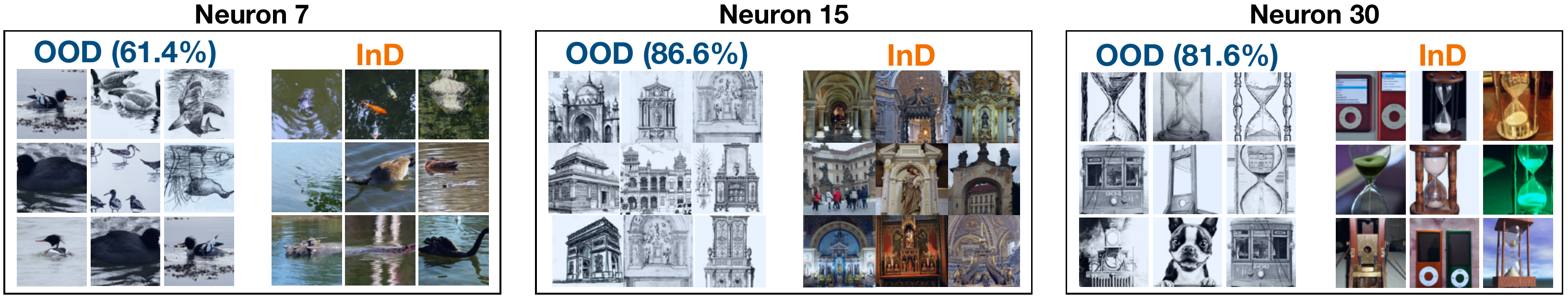} \\
    (a) \\[1em]
    \includegraphics[width=1.0\textwidth]{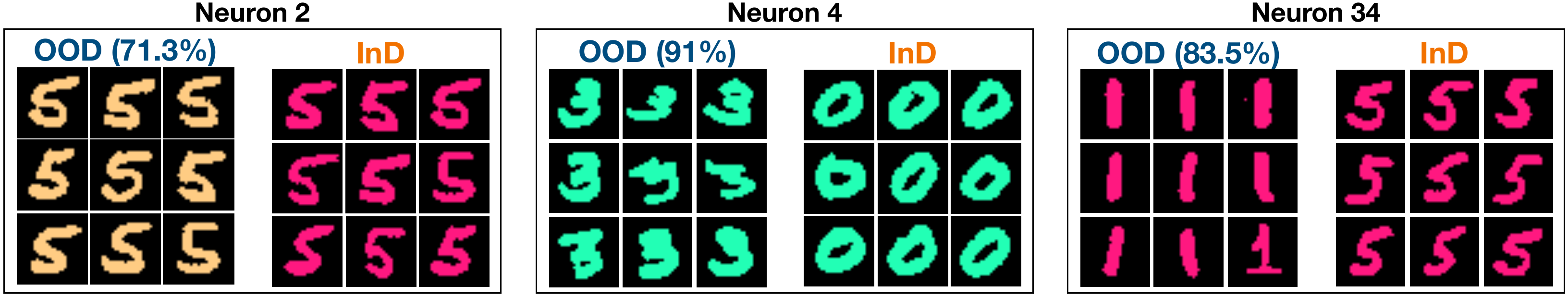} \\
    (b) \\
    \end{tabular}
    \caption{\label{fig:main_image_neuron} \textbf{Feature selectivity and Invariance per neuron.} (a) The \textbf{`Neuron'} view allows visualizing the neural tuning for every neuron along with the highest activation ratio for OOD. Three random neurons (7, 15 and 30) from penultimate layer are visualized here, indicating that they fire for animals or birds in water, ancient buildings or architectures, and hourglass or similar objects, but are invariant to colors. The firing intensity of these neurons are shown by activation ratios 61.4\%, 86.6\% and 81.6\% for OOD compared to InD. (b) Considering random neurons for a different model with Colored MNIST (InD) and Permuted Colored MNIST (OOD), we observe tuning to shape
    of the digit with 71.3\% activation ratio (neuron 2), where as neurons 4, 34 are tuned to spurious feature ’color’ and selective to different categories with even higher activation ratios i.e. 91\% and 83.5\% respectively.}
\end{figure*}

\begin{figure*}[h]
    \centering
    \begin{tabular}{c}
    % \includegraphics[width=1.0\textwidth]{figures/main_images/main_image_1.pdf} \\
    % (a) \\ [1em]
    \includegraphics[width=1.0\textwidth]{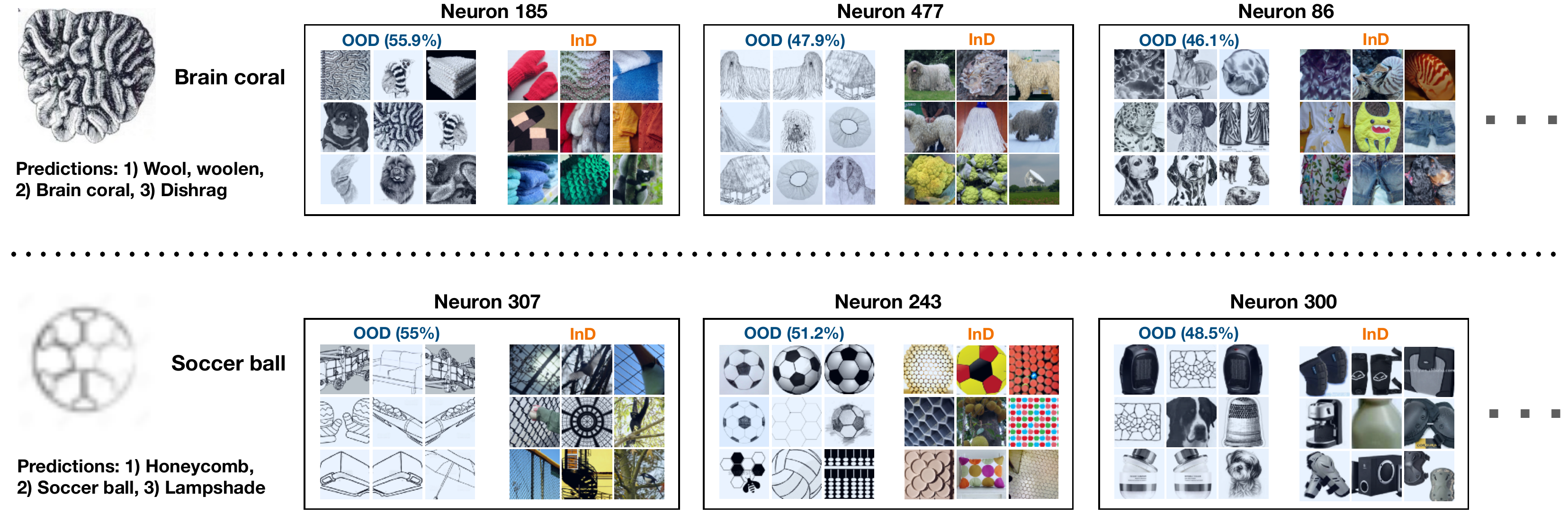} \\
    (a) \\[1em]
    \includegraphics[width=1.0\textwidth]{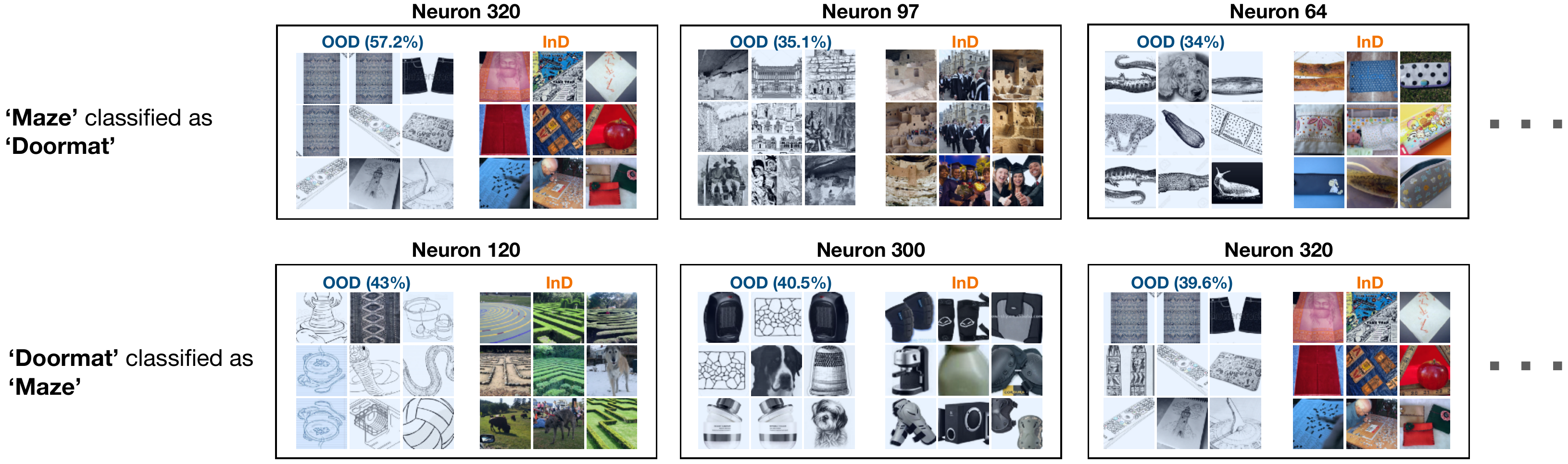} \\
    (b) \\
    \end{tabular}
    \caption{\label{fig:main_image_category} \textbf{(a) Activity per Image.} Any image, either from InD or OOD, can be examined by the \textbf{`Image'} view through a comparative visualization of the most activated neurons for the selected image. Specifically, this helps explaining failure cases from OOD, such as a `Brain coral' image (id 5583) is predicted as `Wool', and the most activated neurons (185, 477, 86) for this image fire for `Wool' category images, as evident from the neural behaviour for InD. Similar explanations for the second `Soccer ball' image (id 40983) are shown with neurons 307, 243 and 300. \textbf{(b) Activity per Category Confusion.} Images from a specific category or a pair of categories, where images from one category are predicted as another category, are investigated in the \textbf{`Category'} view through similar visualization as (a), but by picking the most activated neurons for a set of images under consideration. We have shown examples of `Maze' confused with `Doormat' with top neurons (320, 97, 64) and vice versa (120, 300, 320).}
\end{figure*}

% \begin{figure*}[!h]
% % \vskip 0.2in
% \begin{center}
% \includegraphics[width=180mm,height=185mm]{figures/main_image.pdf}
% \vskip -0.15in
% \caption{\textbf{Explaining Steps of \textit{Deephys}.} }
% \label{fig:main_image}
% \end{center}
% % \vskip -0.2in
% \end{figure*}

\section{Results}
\label{Results}

% \subsection{Deephys: Deep Electrophysiology}
Here we present detailed insights about each step of our proposed methodology and their contributions towards achieving the broader goal of demystifying model failures under distribution shift. 
The steps include analysis of individual neurons to understand feature selectivity and invariance, investigating activity per image as well as a set specifically selected images.
We also explain how \textit{Deephys} can leverage such analyses, generated both for InD and OOD, seamlessly through concept-based visualization technique.
We consider examples from Colored MNIST (added separate color to each MNIST category) and ImageNet, with their OOD versions (Permuted Colored MNIST and ImageNet Sketch respectively) to explain different steps of our approach, that are shown in Fig.\ref{fig:main_image_neuron} and \ref{fig:main_image_category}. Please refer to Sec.\ref{Method} for explanations about the datasets and the networks used for our experimentation.
% from a ResNet18 architecture with ImageNet (InD) and ImageNet Sketch (OOD) (refer to Sec.\ref{Method} for explanations about the datasets) to explain different steps of our approach, that are shown in Fig.\ref{fig:main_image_neuron} and \ref{fig:main_image_category}.
We conclude this section with specific examples from all datasets (including OOD versions), that are used for our experimentation, to show effectiveness of our proposed approach.
It is worth mentioning that investigating any layer of a model can provide important insights about it's behavior, and can be performed through \textit{Deephys}. Yet, we only consider the penultimate layer for this study due to the closest proximity to the decision layer, that may reveal the most differentiating traits at the neuron level.
% We now present specific insights about each step and their contributions towards achieving the broader goal of demystifying model failures under distribution shift.

% \subsection{Neuron View}
\subsection{Visualizing Feature Selectivity by exhibiting most activated images for neurons}
Previous studies have shown that neurons can be considered as feature detectors \cite{elhage2022superposition}, which can be any concept or artifact of the data i.e. color, shape, texture etc. with a specific human understanding. Each image possibly consists of many such concepts and visualizing the top activated images for a neuron provides a sense of the feature selectivity of the neuron.

It is also possible for some neurons to learn superimposed features, leading to interference among them, that make them less or completely uninterpretable for a human observer.
% Each neuron captures one or more concepts, but multiple neurons can represent similar concepts.
Our visualizations also suggest similar behavior as there are some neurons for which the top activated images may not seemingly share any particular feature. Moreover, while some features of a complex natural dataset can not be easily named, they may be intuitively picked up by looking at the top activated images. Here we allow the human observer to visualize the neuron's response patterns and interpret what patterns require a closer examination.

% This toolset is about exhibiting the top activated images for a neuron, from the dataset used for training the model, hence considered as visualization with InD. 
The `InD' heads of Fig.\ref{fig:main_image_neuron}(a) show the most activated images for 3 sample neurons, from the penultimate layer of a ResNet18 model, for \textbf{InD} (i.e. ImageNet).
They demonstrate the selectivity of the neurons to features such as animals or birds in water (neuron 7), ancient buildings or architectures (neuron 15), and hourglass or similar looking objects (neuron 30). 
We add other examples for Colored MNIST (separate colors to each MNIST category) as \textbf{InD} and Permuted Colored MNIST (keep same colors and change the color-category association by random permutation of colors) as \textbf{OOD} for better explanation. More details about different versions of Colored MNIST are provided in Sec.\ref{Method}. The `InD' heads of Fig.\ref{fig:main_image_neuron}(b) exhibits that the 
neurons are tuned to pink `5' (neurons 2, 34) or green `0' (neuron 4).

% Fig.\ref{main_selectivity} shows examples of Deephys displaying top 100 images from ImageNet for two sample neurons from the penultimate layer of a ResNet18 model trained on ImageNet. We name the penultimate layer as '\textit{linear1}' and is simply a naming convention we used. We can generate such visualization for any neuron in the selected layer.

\subsection{Visualizing Invariance by comparing most activated Images for InD and OoD}
While visualizing neural responses to InD data is beneficial towards understanding feature selectivity, following the same procedure for OoD can enlighten us about invariance or the artifacts, that do not affect the neurons to respond.
Therefore, generating comparative visualization of the neurons, both for InD and OoD, based on the same concept-based explainability technique can lead to in-depth exploration of invariance through different bias factors of the datasets.
% and interpretation of model's invariance and failure against OoD.
Our proposed method can produce such synchronous visualization for any neuron from the model and allows the observer to investigate their behavioral change and capture a broad understanding about their invariance.
% their characteristics individually.
% Comparing the behaviour of the neurons for InD with multiple OoD datasets simultaneously can be more illuminating in terms of their invariance. This convinced us to add the feature of visualizing neurons for multiple OoD datasets in Deephys at the same time.

As the neurons are tuned to patterns originating from InD data, they respond on encountering similar patterns in OOD data. 
The activation ratio of a neuron for OOD data compared to InD data  (considering only the highest activation scores for both the datasets) can reflect the similarity of features.
Also, some neurons can be tuned to spurious features in InD data. When the spurious features are associated to different categories in OOD data, the neurons can become selective to these categories. These phenomenon can be observed by our idea of comparative visualization of neurons.  

Fig.\ref{fig:main_image_neuron}(a) exhibits the most activated images, for the same neurons as presented in the last section, for \textbf{OOD} (i.e. ImageNet Sketch).
This shows that the neurons are selective to very similar artifacts from OOD data as in InD data i.e. birds in water (neuron 7), ancient buildings (neuron 15), and hourglass or similar objects or animal body parts (neuron 30), but are invariant to colors. The neural activation ratio for each neuron (considering only the highest activations of that neuron for OOD compared to InD) represents existence of a feature in OOD data with respect to InD data. Here, neurons 15 and 30 show high feature similarity (86.6\% and 81.6\% respectively), but neuron 7 observes less similarity (61.4\%) between OOD and InD.
% Here the neurons fire 61.4\%, 86.6\% and 81.6\% at the maximum for neurons 7, 15 and 30 respectively.
Also Fig.\ref{fig:main_image_neuron}(b) shows that neuron 2 is selective to shape of the digit with 71.3\% activation ratio, where as neurons 4, 34 are tuned to spurious feature `color' and selective to different categories with even higher activation ratios i.e. 91\% and 83.5\% respectively.

% The fig.\ref{fig:main_image_neuron} exhibits the most activated images for the same 3 sample neurons for \textbf{OoD} (i.e. ImageNet Sketch), which were selected for InD. This shows that the selected neurons fire for very similar artifacts from OoD as in the InD, even though the colors of the objects are missing in OoD. In other words, these 3 neurons are mostly selective to specific characteristics, but are invariant to colours.
% By exploring the comparative visualization for different neurons both from InD and OoD efficiently through deephys, the user may have a better comprehension of the tuning of neurons, compared to when only analyzing the neurons for InD.

\begin{figure*}[h]
    \centering
    \begin{tabular}{cc}
    \includegraphics[width=0.18\textwidth]{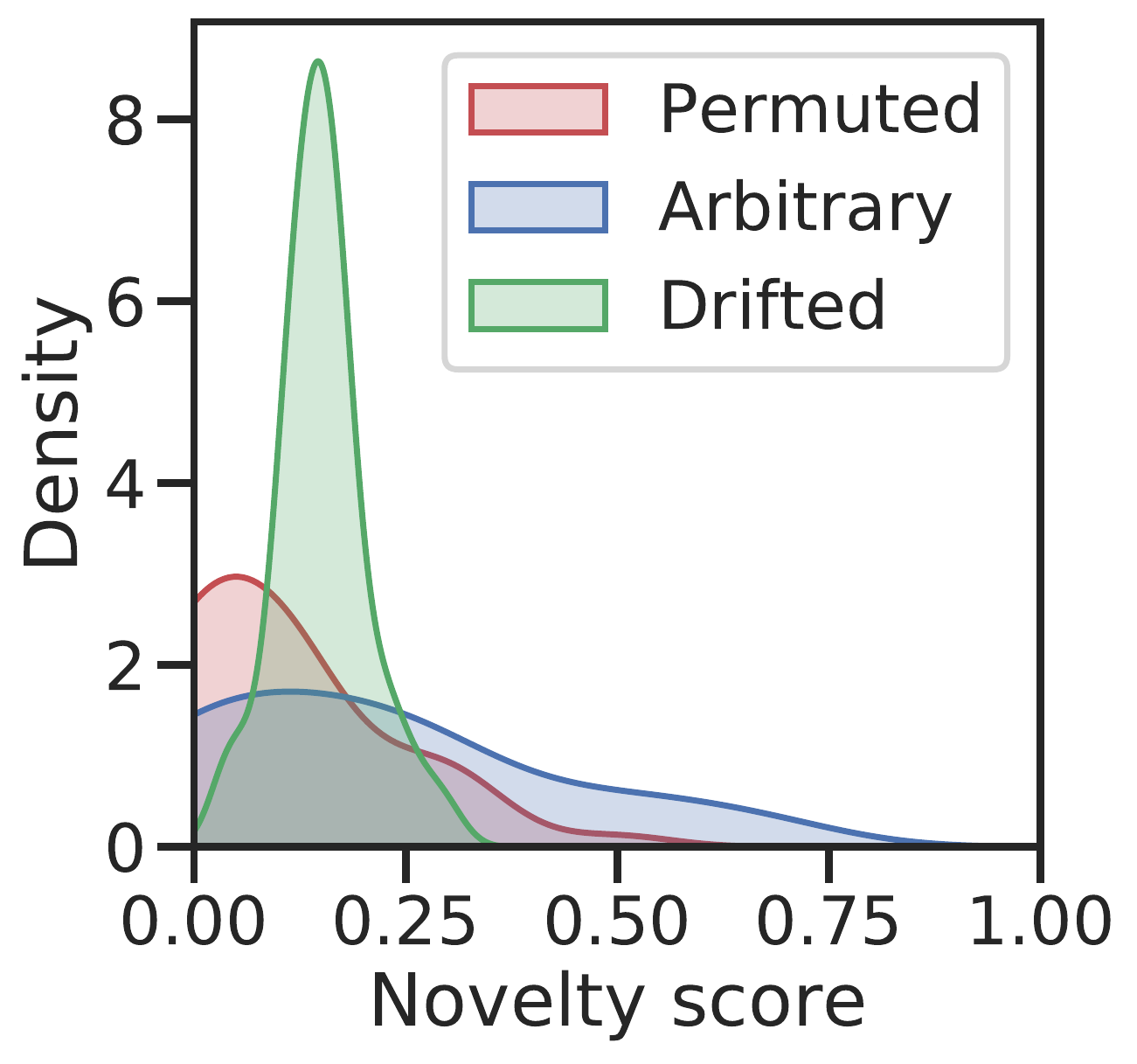} &
    \includegraphics[width=0.6\textwidth]{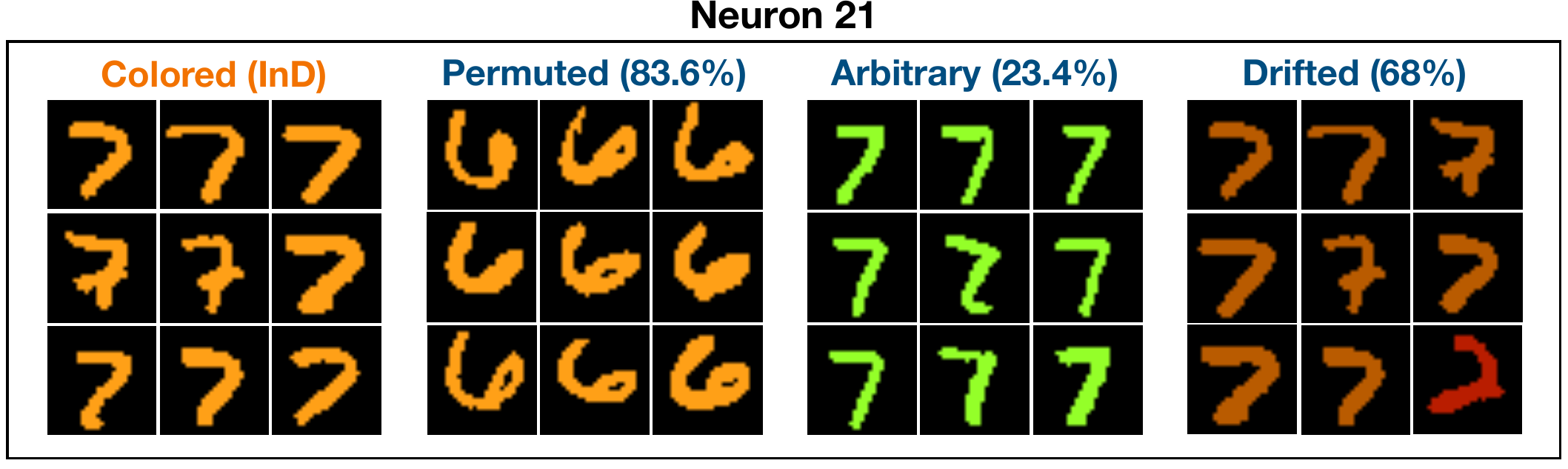} \\
    (a) & (b)\\ [1em]
    \includegraphics[width=0.2\textwidth]{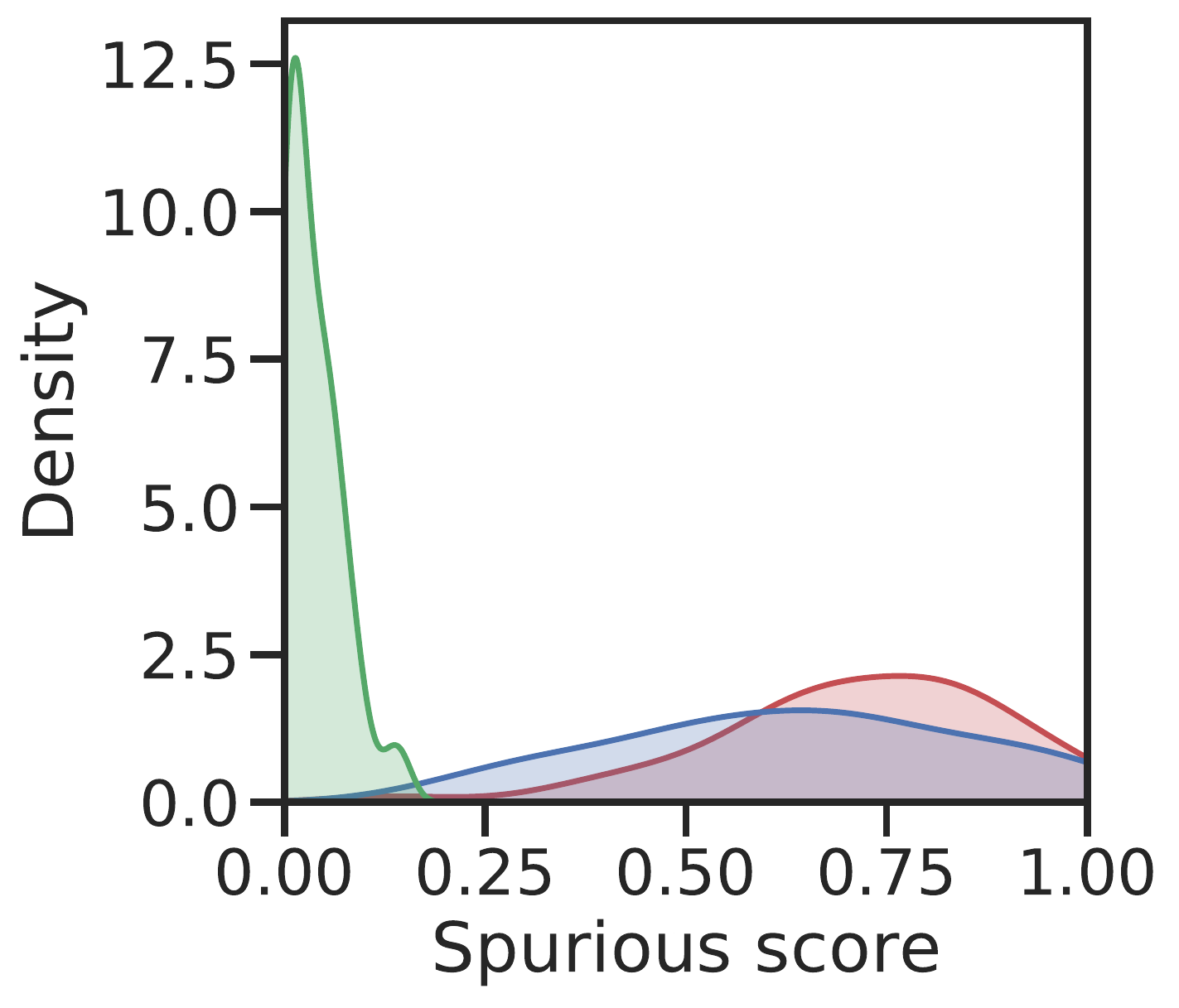} &
    \includegraphics[width=0.6\textwidth]{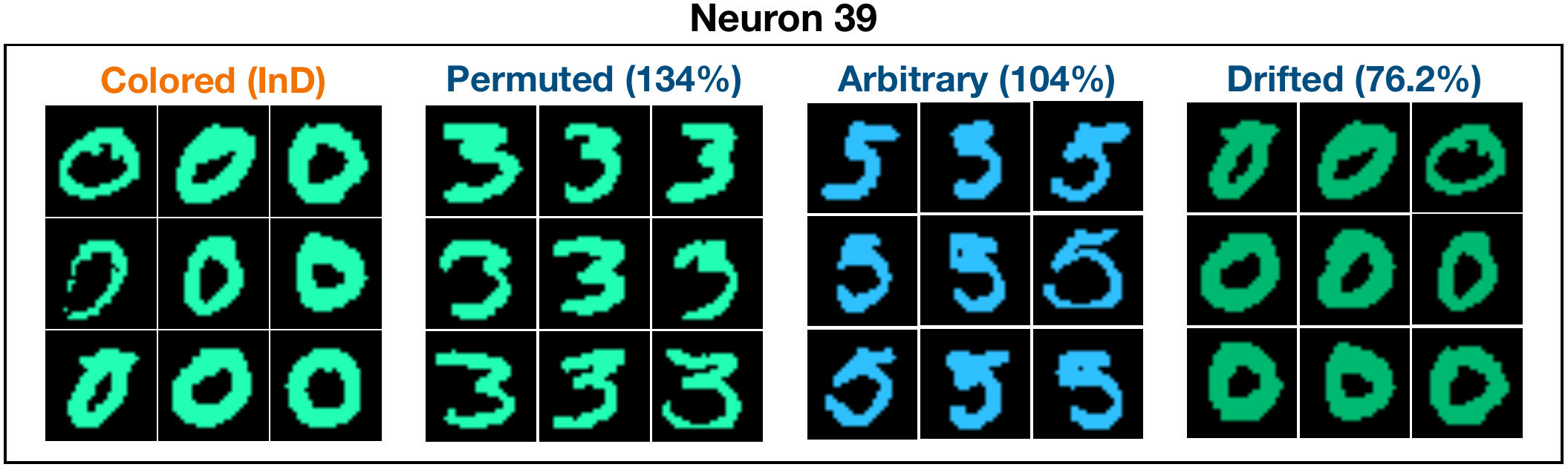} \\
    (c) & (d)\\ [1em]
    \end{tabular}
    \begin{tabular}{c}
    \includegraphics[width=0.96\textwidth]{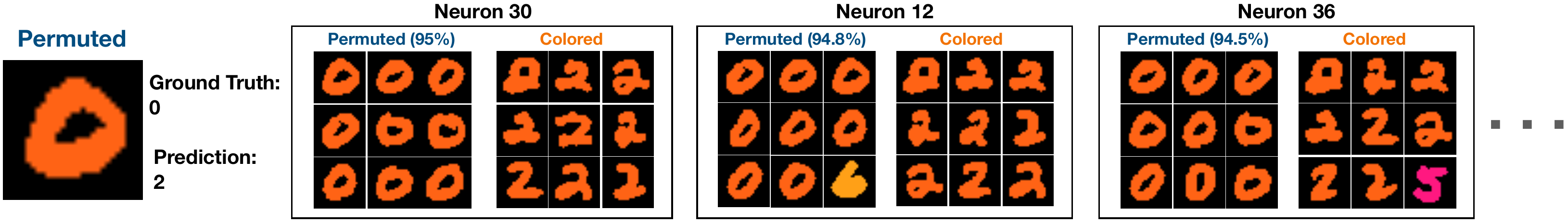} \\
    (e) \\ [1em]
    \includegraphics[width=0.96\textwidth]{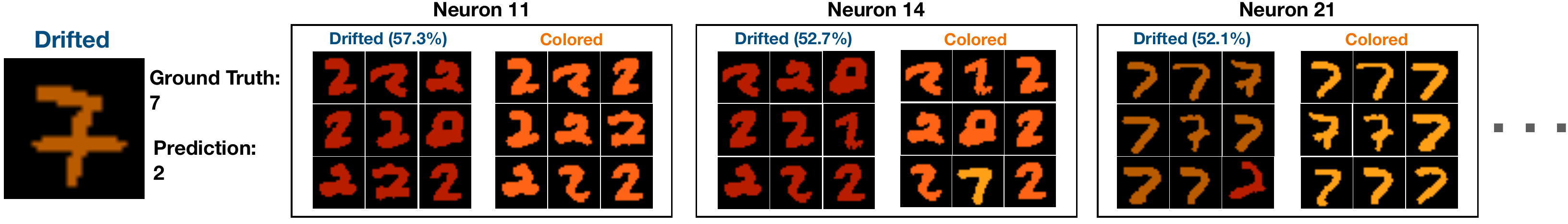} \\
    (f) \\
    \end{tabular}
    \caption[XYZ]{\label{fig:all_MNIST} Analysis of (a) new features and (c) spurious features in different OOD version of `Colored MNIST' (InD) i.e. `Permuted Colored MNIST', `Arbitrary Colored MNIST' and `Noisy Colored MNIST'. One example is considered representing each neural property in OOD i.e. new features ((b) neuron 21) and spurious features ((d) neuron 39). Failure cases from OOD versions are investigated with compartive visualization of most activated neurons of OOD as well as InD ((e) neurons 30, 12, 36), and (f) neurons 11, 14, 21).}
\end{figure*}

\subsection{Analyzing Activity per Image}
In addition to investigating the behaviour of individual neurons, it is also essential to understand the responses of them for individual images in different datasets.
The neural responses are normalized for every neuron (i.e. dividing the activations of all the images by the highest score for that neuron) for InD, that brings all the normalized activations between 0 and 1. Please note that we consider only positive neural activations (by applying \textit{ReLu}) before applying a normalization.
For a selected image, the neurons are identified by their descending normalized activations, that are mostly activated for the image. Such normalized scores are attached with the neurons, that reflects their individual importance for the image. For complete understanding, we follow the same concept-based approach of explaining a single neuron as explained for the previous steps, and equip every neuron in this `per \textbf{image}' view with such visualizations.
% We show neurons of a particular layer that are mostly activated by a specific image, sorted by their descending firing rates for the selected image. These scores of every neuron for the selected image are also displayed for further investigation regarding relative importance of each neuron. Following the same approach of visualizing a single neuron as the previous steps of Deephys, we show top activated images for every neuron in addition to the scores. Such visualizations can provide instant insights into the response behavior of individual neurons with regard to a specific image. 
Such visualizations can provide insights about perception of a model for an image in terms of interpreting individual neurons that are important for the image.
Generally, one or a few of the top most activated neurons are expected to be similarly tuned to the features of the selected image. These top neurons can be individually tuned to different patterns of the image based on the complexity and coexistence of dissimilar features in the image.

As the purpose of this study is to compare model behavior for OOD to InD, we follow the same normalization procedure here through dividing all the activations of OOD images of a neuron by the highest score of that neuron for InD (i.e. neural response ratio). For a selected InD image, we include visualizations of the same set of neurons as InD, with the neural response ratio and the top activated images from OOD for every neuron.
On the contrary, it is also possible to select an image from OOD and generate a similar comparative visualization. We always calculate the normalization with respect to InD, but here the highest activated neurons are identified for the OOD image accompanied by the concept-based visualization, and the same order of neurons are followed for such visualizations with InD.
% Also, we can select any image from OoD and generate a similar comparative by-image visualization. Just as before but reversed, the top images are now calculated as the highest activated neurons to the OoD image, and the InD dataset shows coresponding neurons in the same order. 
Such visualization demonstrates overall behaviour of a model by explaining individual images from OOD and helps to identify inconsistent performance through explaining the change in response of most important neurons for the selected image.
% This can also lead to explanations for unexpected decisions taken by the model for OoD images.

Part (a) of Fig.\ref{fig:main_image_category} exhibits the capability of \textit{Deephys} to select an image from OOD and visualize the corresponding top activated neurons while comparing their behavior of the same set of neurons from InD. Following the interest in understanding failure cases from OOD, two sample images are considered for analysis from ImageNet Sketch (i.e. OOD). For the ‘Brain coral’ image (id 5583), predicted as `Wool' by the model, the most activated neurons (i.e. 185, 477, 86,..) are selective to features seemingly similar to `Wool', as evident from visualizations of the same neurons for InD.
Similarly, for the ‘Soccer ball’ image (id 40983), predicted as `Honeycomb', the most activated neurons (i.e. 307, 243 and 300) are tuned to features like honeycomb.
% first image, the model put highest confidence in the category \textit{Wool woolen}, more than the ground truth category \textit{Brain coral}, where as \textit{Dishrag} being the third prediction. Carefully observing the visualization of the top neurons with images from the same dataset (i.e. OoD) and the matching neurons with images from InD justifies the model decision to a great extent, as these neurons are largely selective to woolen objects in this case. A very similar explanation can be derived for the second image with the ground truth \textit{Volleyball}. 

\subsection{Analyzing Category-wise Activity}
% \subsection{Activity per Category Confusion}
Explaining the unreliable behaviour of the model against OOD data may also require interpreting model responses for a set of specific set of images. We consider either all the images of a category or the set of images of a category, that confuse the model to predict another category, to be of great interest towards our goal. 
Such visualization is generated through the similar calculation as in the previous step, but the top neurons are calculated considering all the images of the selected set. Here, the neuron activations are averaged for all the images and presented in a descending order.
% Such a visualization is very similar to finding the most activated neurons and showing top images for these neurons, apart from how  the top most activated neurons are calculated. For every set of images that confuse the model about two specific categories, the neuron activations for the images of the set are averaged and then the neurons are shown with the average scores in a descending order. Finally these neurons are visualized with the average score and top activated images for every neuron to provide a thorough explanation.

\begin{figure*}[h]
    \centering
    \begin{tabular}{cccc}
    \includegraphics[height=25mm]{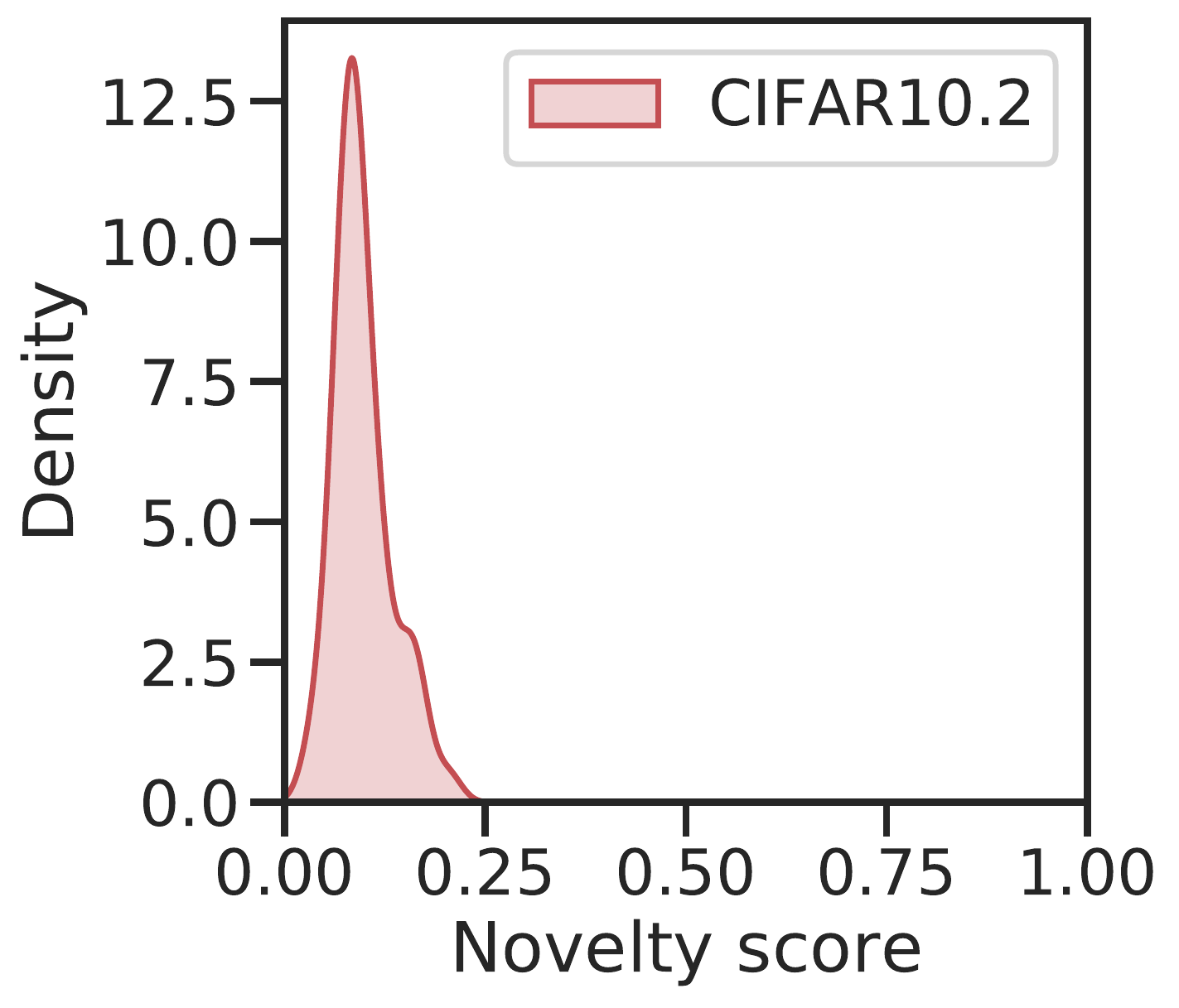} &
    \includegraphics[width=45mm,height=25mm]{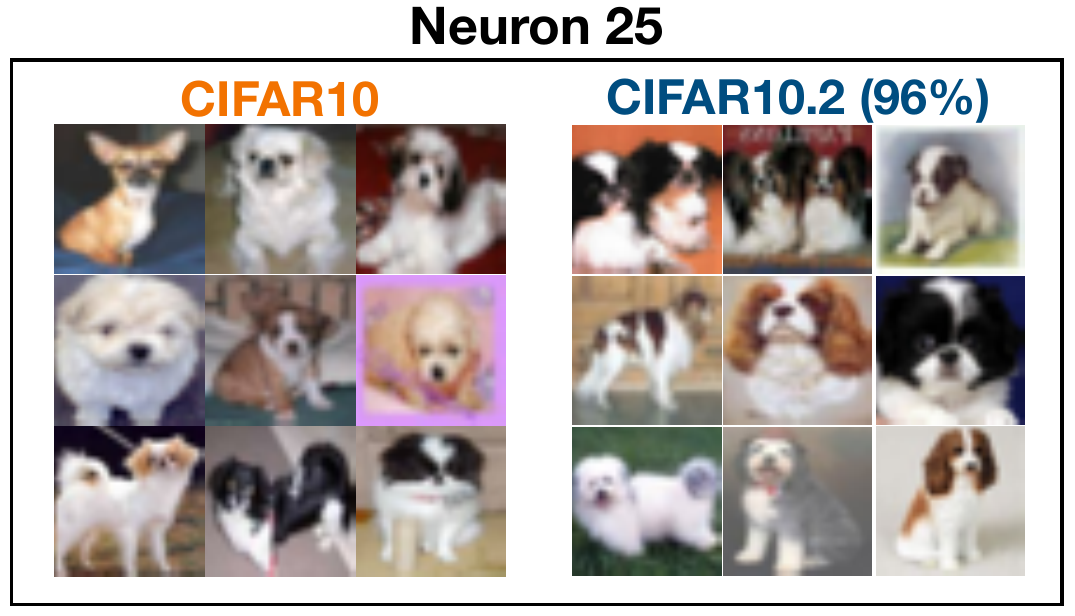} &
    \includegraphics[height=25mm]{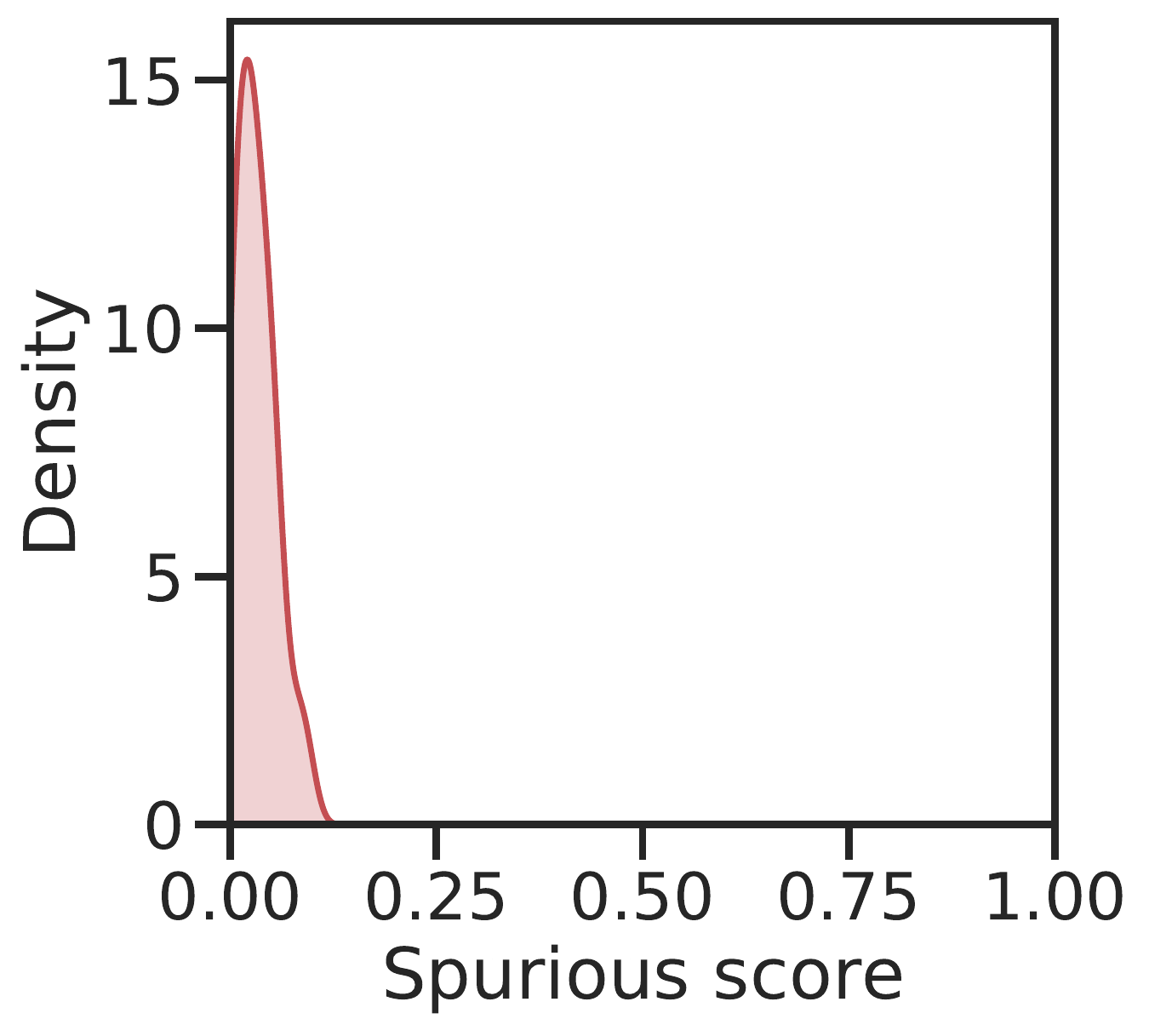} &
    \includegraphics[width=45mm,height=25mm]{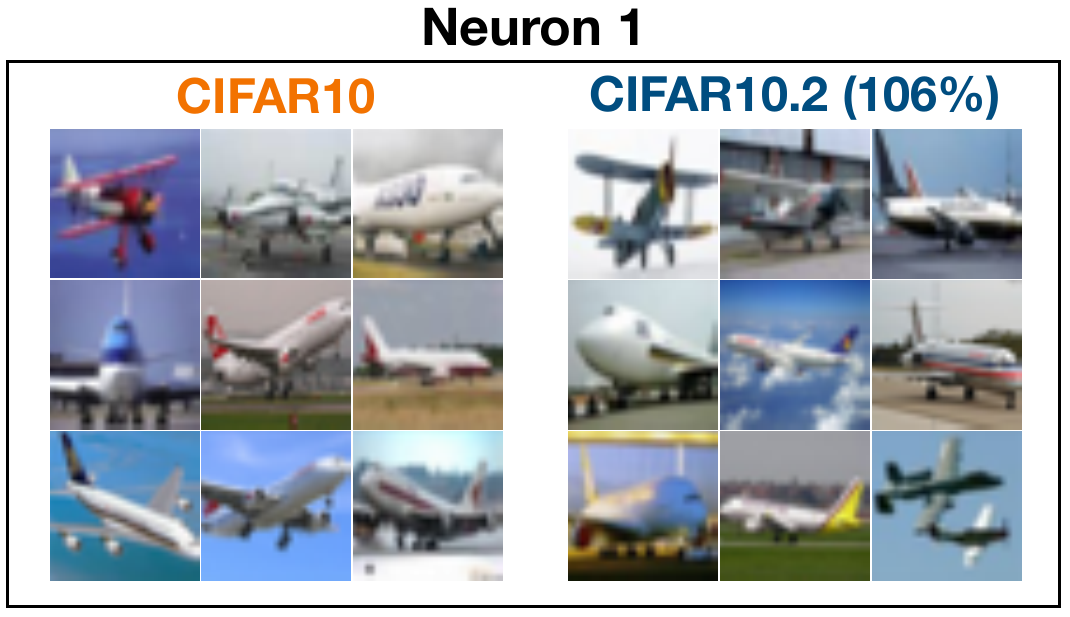} \\
    (a) & (b) & (c) & (d) \\ [1em]
    \end{tabular}
    \begin{tabular}{c}
    \includegraphics[width=\textwidth]{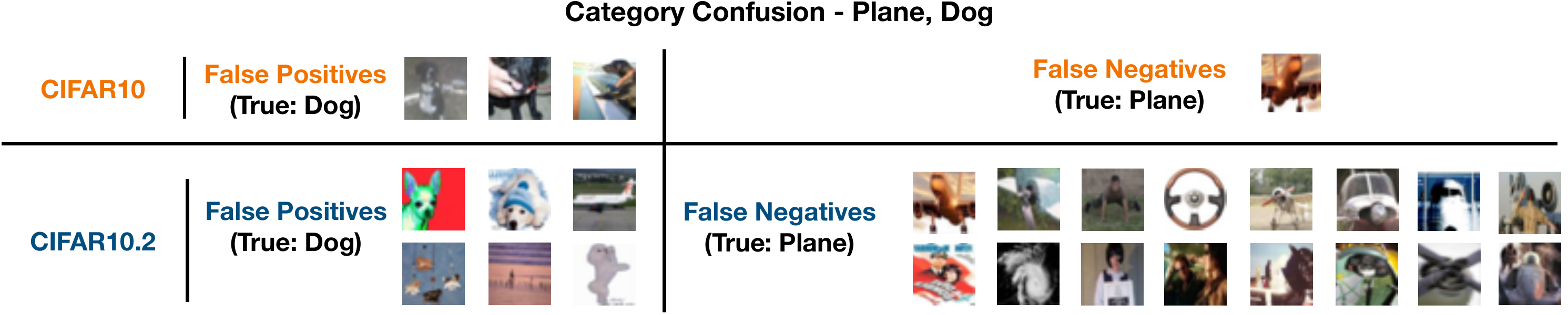} \\
    (e) \\
    \end{tabular}
    \caption[XYZ]{\label{fig:all_CIFAR} Analysis of (a) new features and (c) spurious features in the OOD version of `CIFAR10' (InD) i.e. `CIFAR10.2'. One example is considered representing each neural property in OOD i.e. new features ((b) neuron 25) and spurious features ((d) neuron 1). (e) Confusing images are observed for categories `Plane' and `Dog', both for InD and OOD, to discover multiple cases of mislabelling and difficilut images in OOD compared to InD.}
\end{figure*}

Part (b) of fig.\ref{fig:main_image_category} explains similar visualization with a set of images considered in this part instead of to a single image from OOD. More specifically, the set of images are confusing for a pair of categories, which is of particular interest towards understanding overall model failure. In order for an image to qualify as being part of the confusion between two categories, it must part of at least one instance of a false positive or false negative miss-classification with the other image. Here, analyzing the top neurons for a set of images, with ground truth \textit{Maze} and model prediction \textit{Doormat} holding together, can explain the reason behind such actions by the model, as these neurons are mostly tuned to `Doormat' or objects with very similar attributes.

\begin{figure*}[h]
    \centering
    \begin{tabular}{ccc}
    \includegraphics[width=0.18\textwidth]{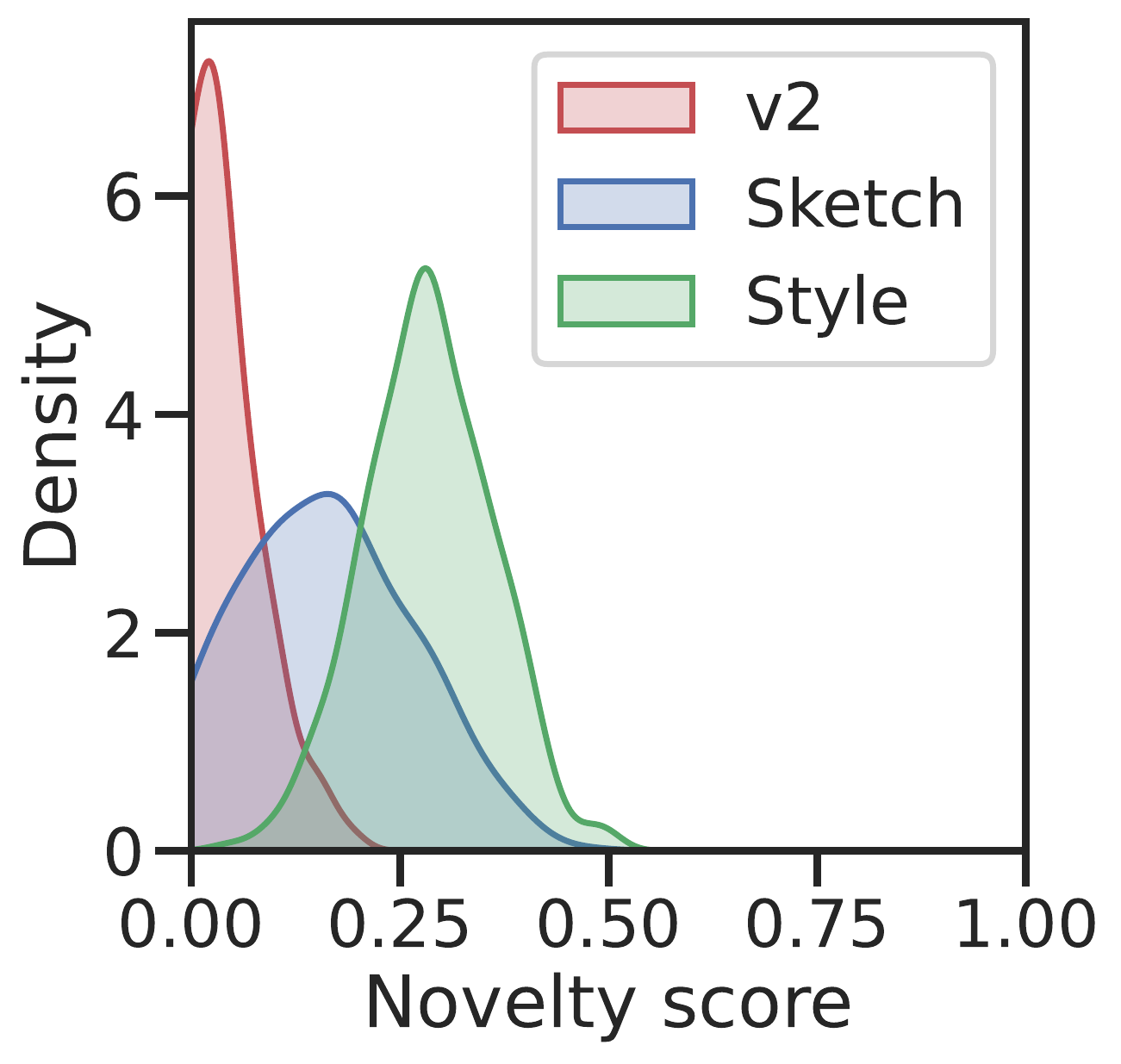} &
    \includegraphics[width=0.6\textwidth]{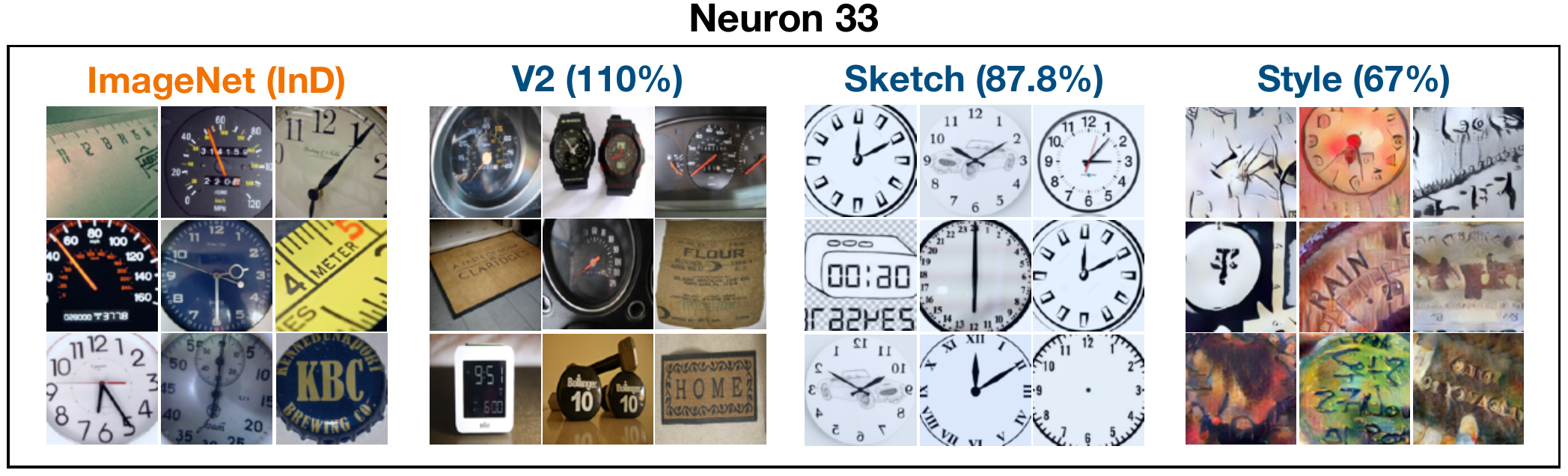} \\
    (a) & (b) \\ [1em]
    \includegraphics[width=0.18\textwidth]{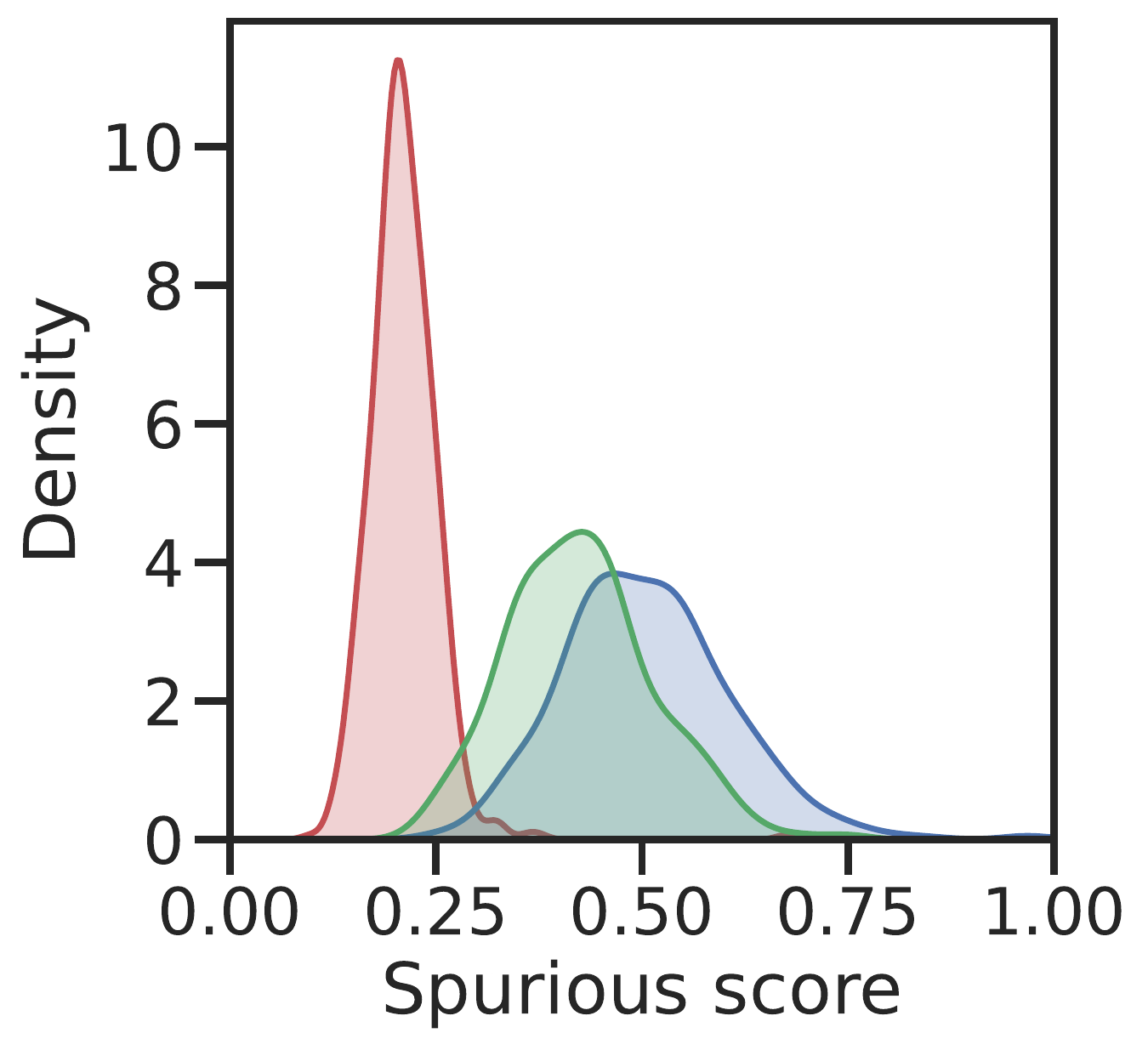} &
    \includegraphics[width=0.6\textwidth]{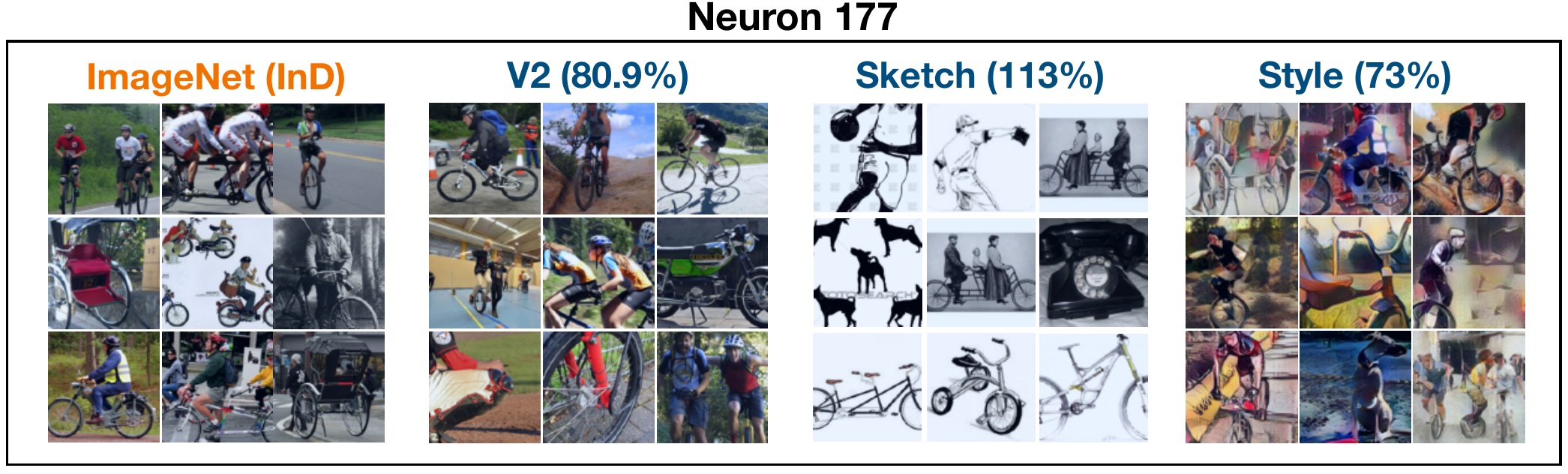} \\
    (c) & (d) \\ [1em]
    \end{tabular}
    \begin{tabular}{c}
    \includegraphics[width=\textwidth]{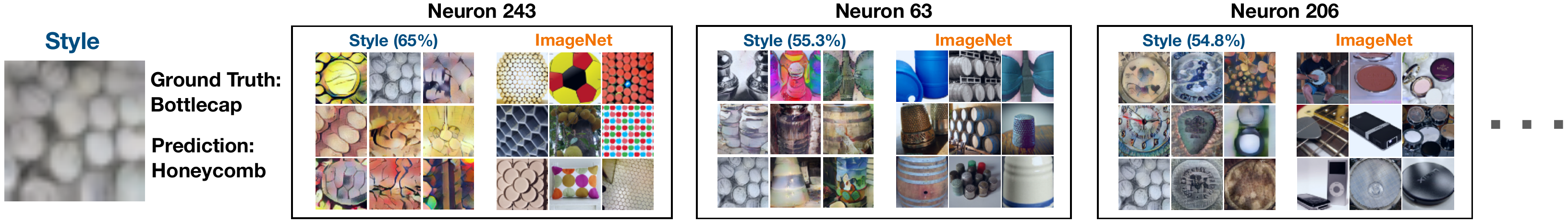} \\
    (e) \\
    \end{tabular}
    \caption[XYZ]{\label{fig:all_IN} Analysis of (a) new features and (c) spurious features in different OOD version of `ImageNet' (InD) i.e. `ImageNetV2', `ImageNet Sketch' and `Stylized ImageNet'. One example is considered representing each neural property in OOD i.e. new features ((b) neuron 33) and spurious features ((d) neuron 177). One failure case from `Stylized ImageNet' is investigated with comparative visualization of most activated neurons of OOD as well as InD ((e) neurons 243, 63, 206). All these visualizations are generated with a ResNet18 model trained on ImageNet (InD).}
\end{figure*}

\begin{figure*}[h]
    \centering
    \begin{tabular}{cccc}
    \includegraphics[width=0.2\textwidth]{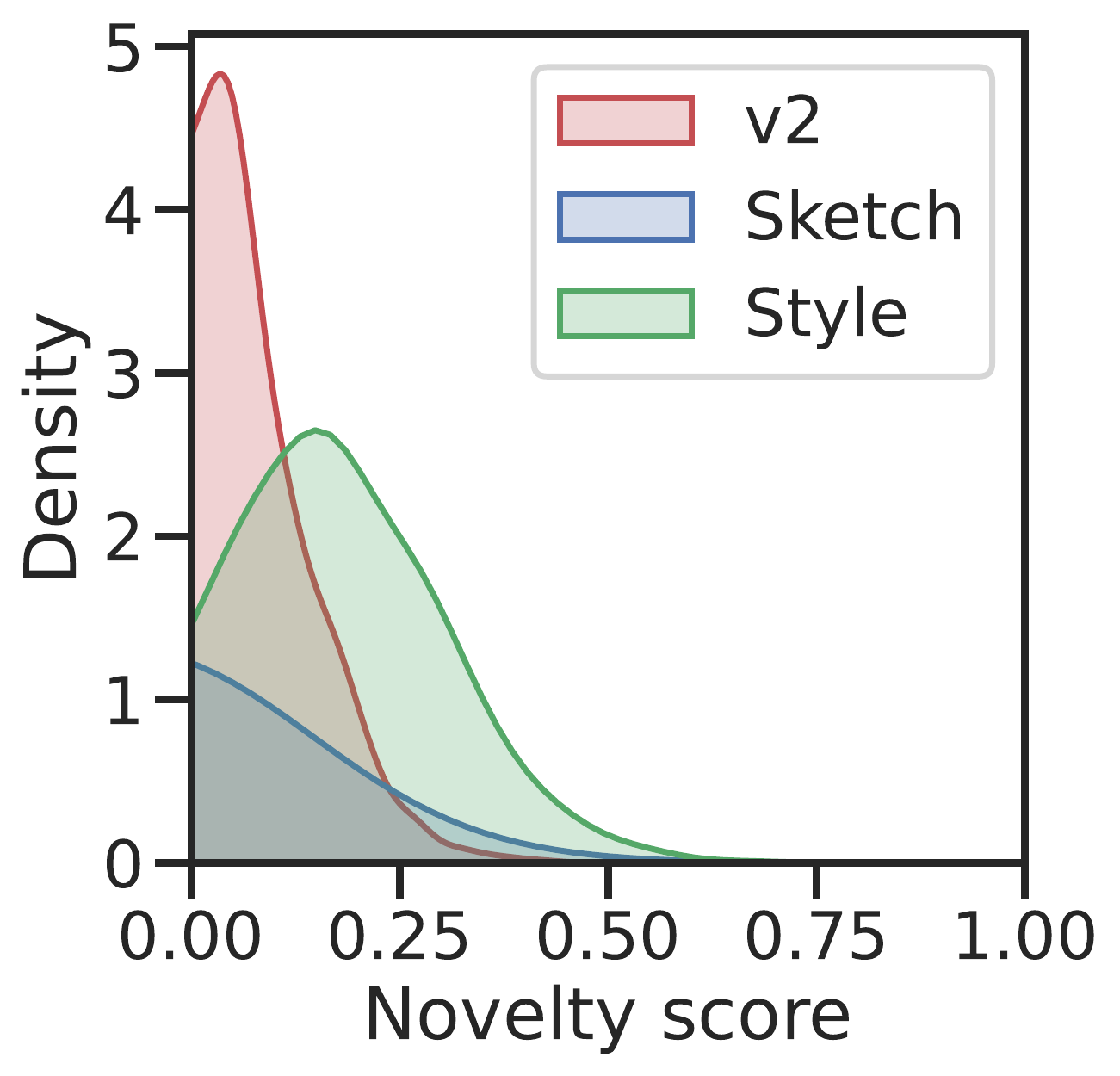} &
    \includegraphics[width=0.2\textwidth]{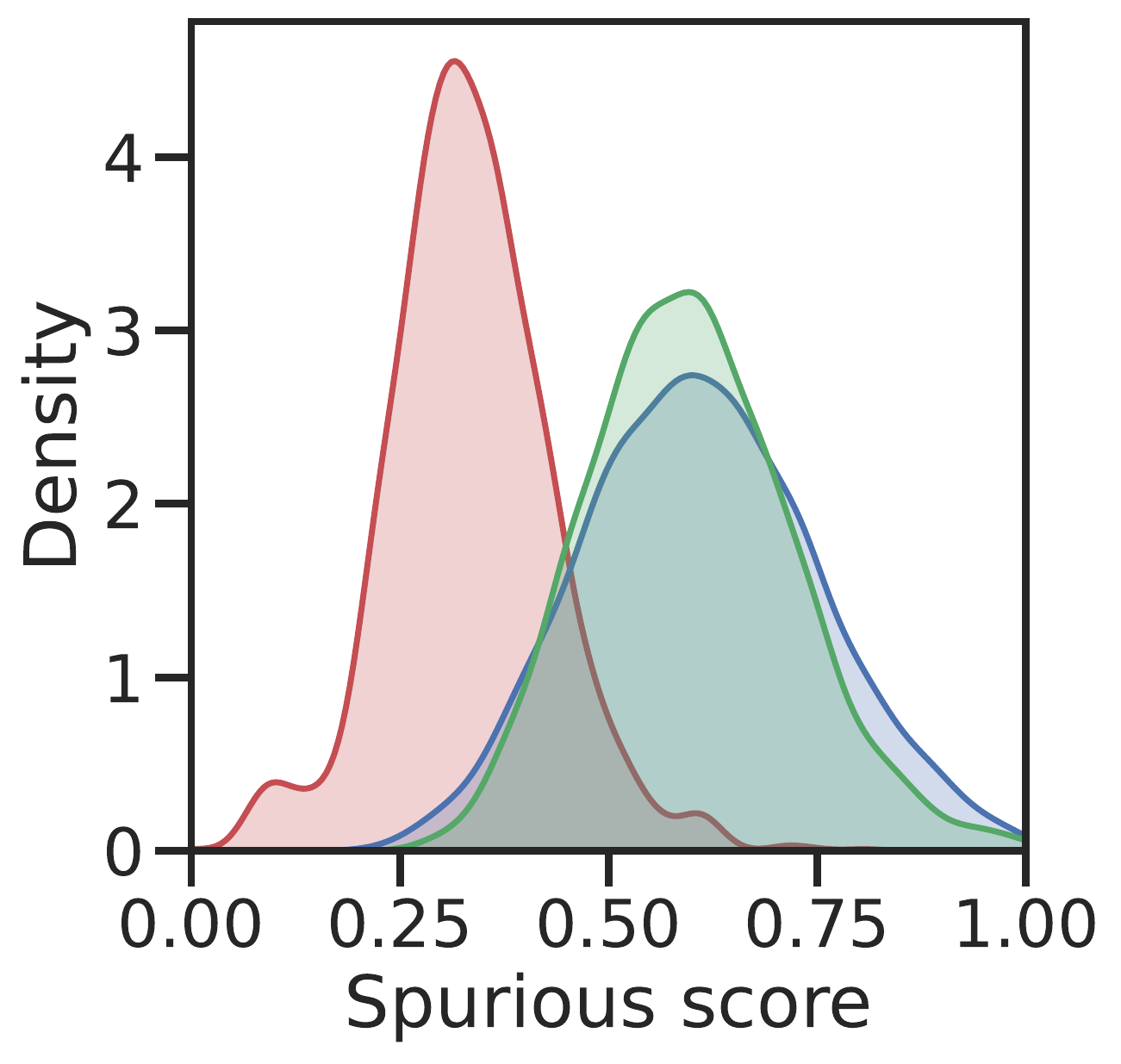} &
    \includegraphics[width=0.21\textwidth]{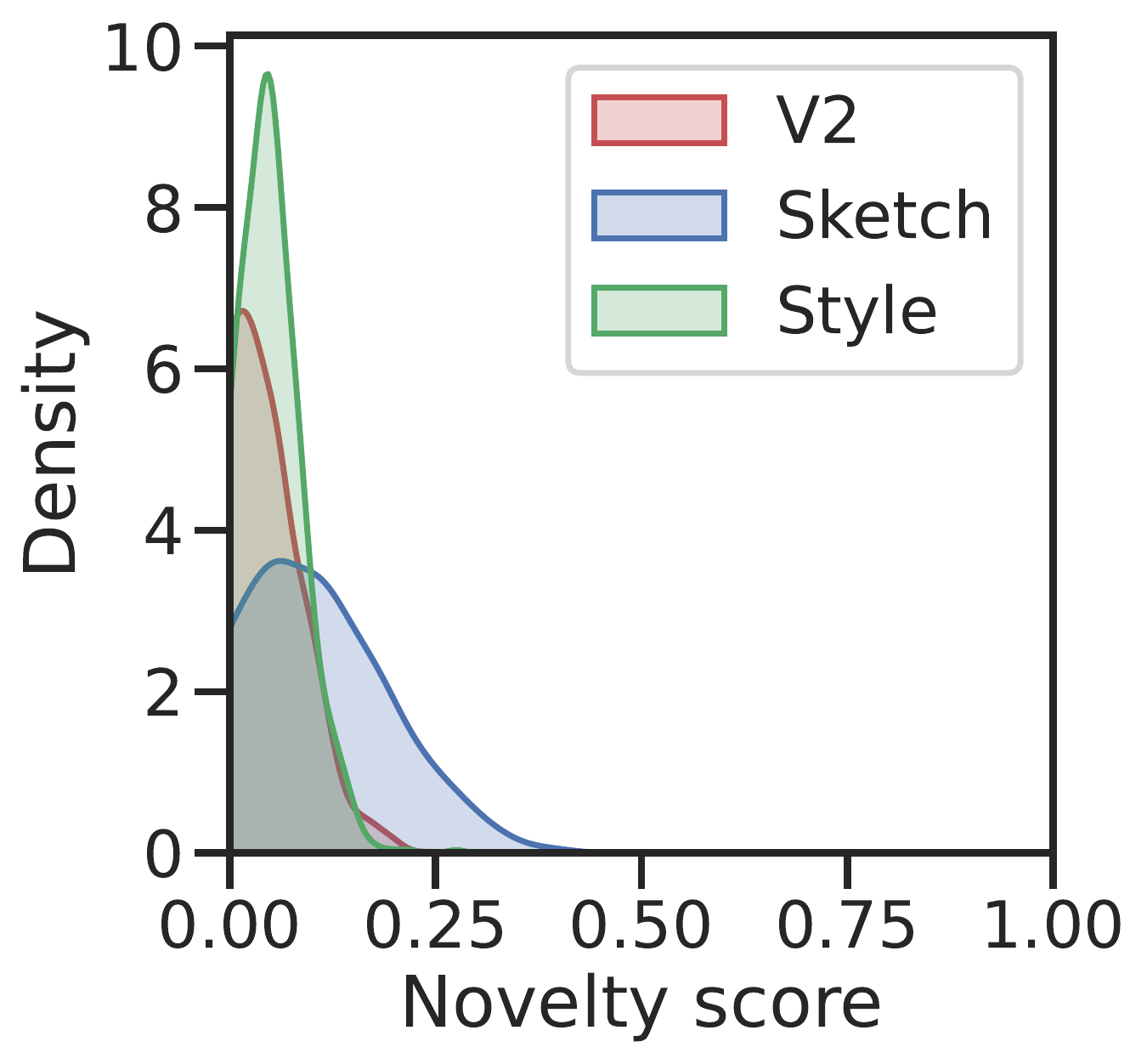} &
    \includegraphics[width=0.21\textwidth]{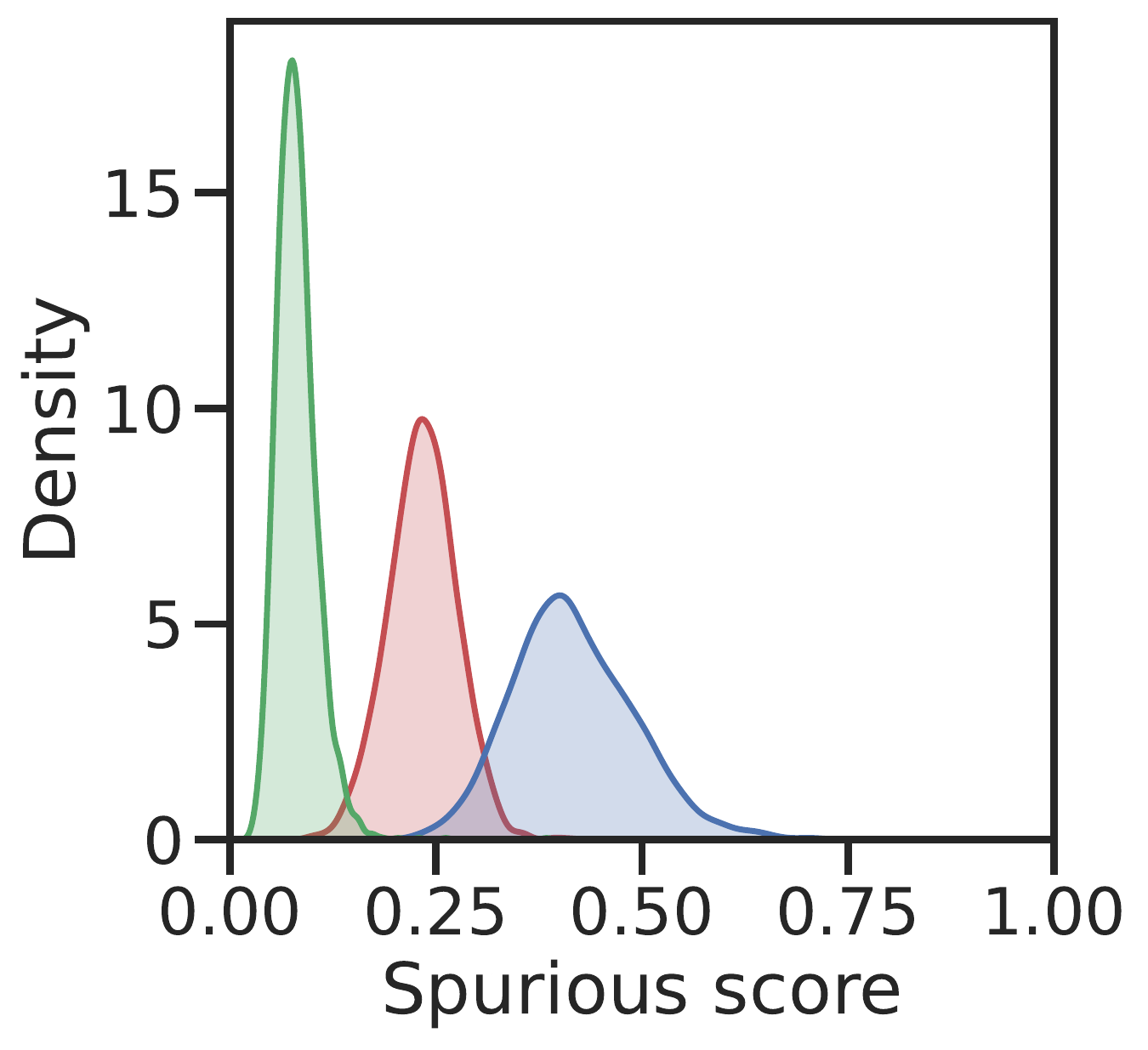} \\
    (a) & (b) & (c) & (d)
    \end{tabular}
    \caption[XYZ]{\label{fig:all_IN_R50} Analysis of (a) new features and (b) spurious features in different OOD version of `ImageNet' (InD) i.e. `ImageNetV2', `ImageNet Sketch' and `Stylized ImageNet', where the model is ResNet50. (c) and (d) show the similar analysis where the ResNet50 network is trained on `Stylized ImageNet'.}
\end{figure*}

\begin{figure*}[h]
    \centering
    \begin{tabular}{ccc}
    \includegraphics[width=0.18\textwidth]{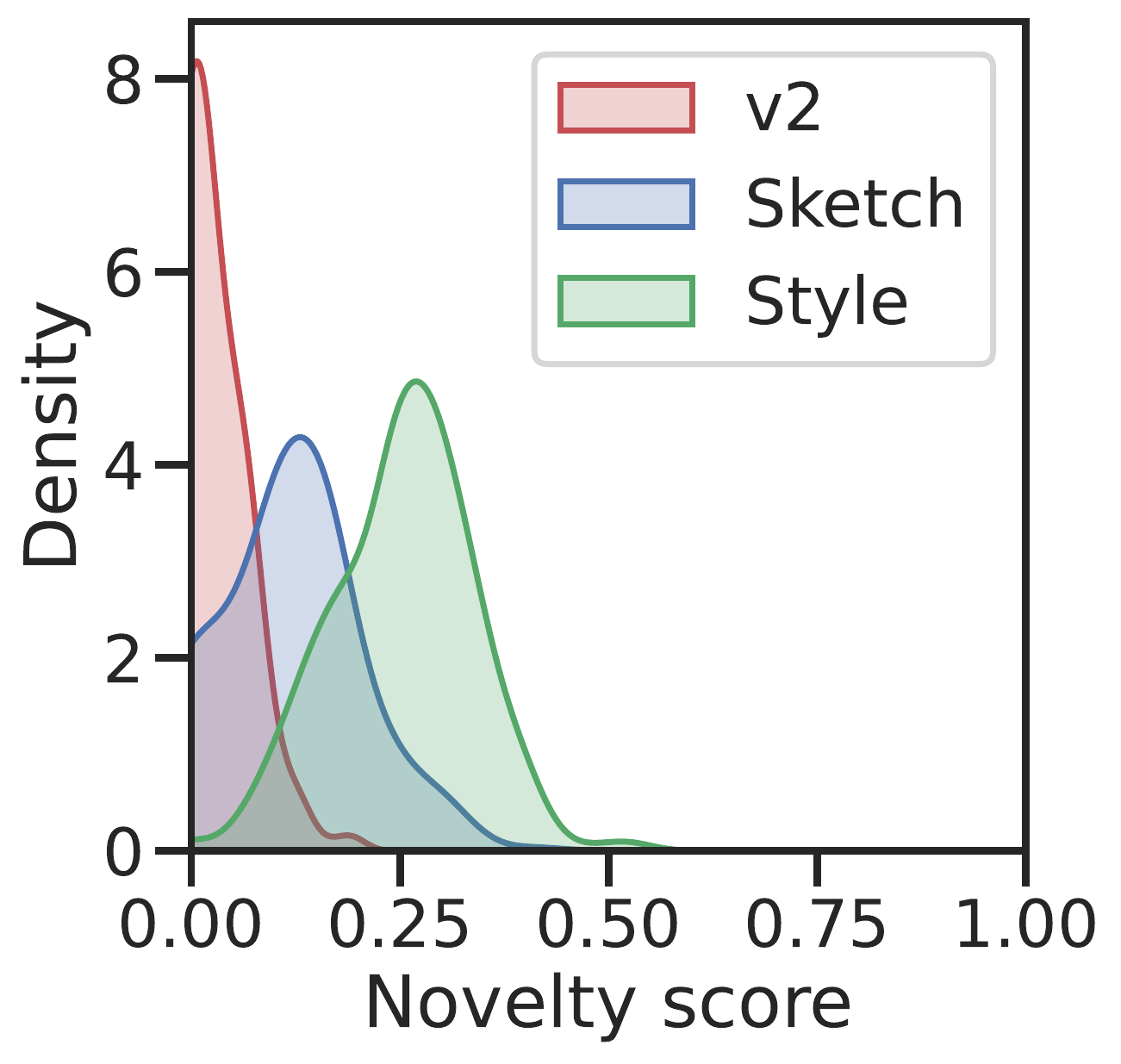} &
    \includegraphics[width=0.6\textwidth]{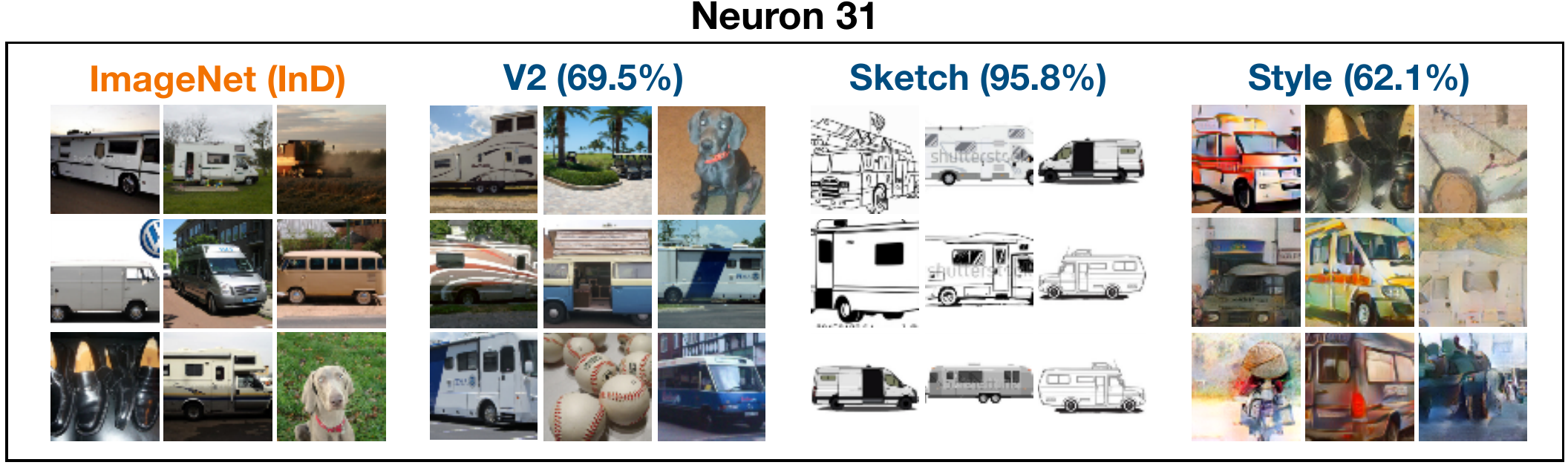} \\
    (a) & (b) \\ [1em]
    \includegraphics[width=0.18\textwidth]{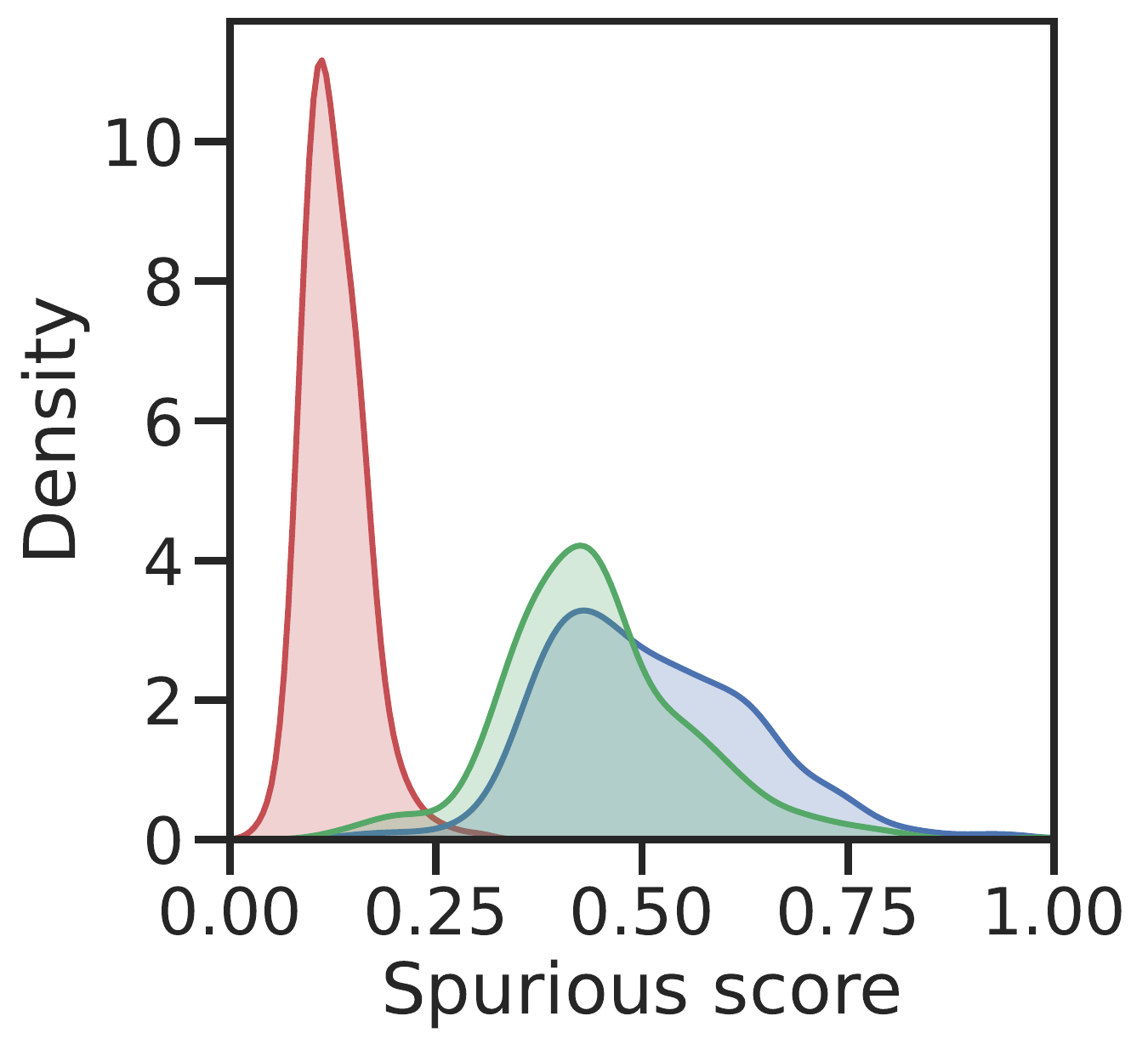} &
    \includegraphics[width=0.6\textwidth]{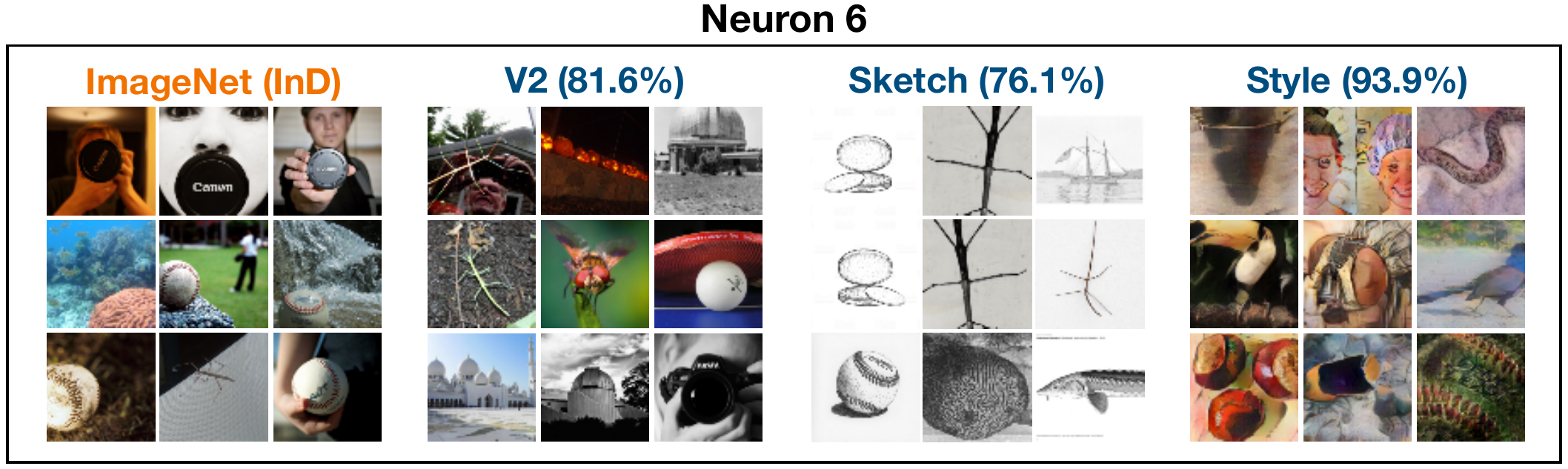} \\
    (c) & (d) \\ [1em]
    \end{tabular}
    \begin{tabular}{c}
     \includegraphics[width=\textwidth]{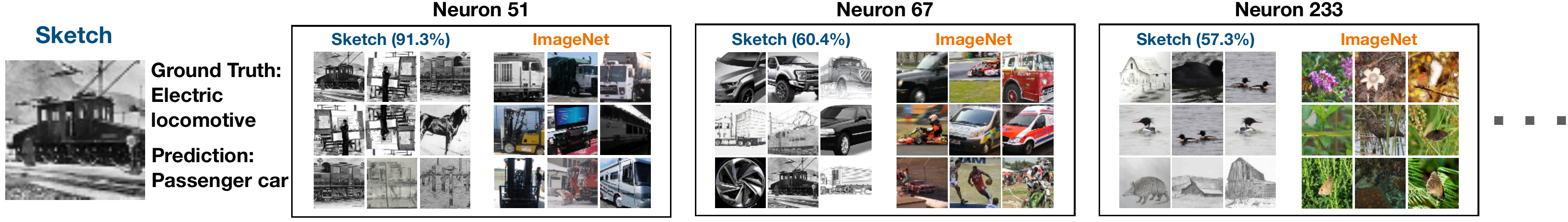} \\
     (e) \\
    \end{tabular}
    \caption[XYZ]{\label{fig:all_IN_CVT} Analysis of (a) new features and (c) spurious features in different OOD version of `ImageNet' (InD) i.e. `ImageNetV2', `ImageNet Sketch' and `Stylized ImageNet'. One example is considered representing each neural property in OOD i.e. new features ((b) neuron 31) and spurious features ((d) neuron 6). One failure case from `Stylized ImageNet' is investigated with comparative visualization of most activated neurons of OOD as well as InD ((e) neurons 51, 67, 233). All these visualizations are generated with a Convolutional Vision Transformer (CvT) model trained on ImageNet (InD).}
\end{figure*}

\subsection{Designing Deephys Software}

As a single biological neuron's firing properties in response to different stimuli are rigorously compared in an electrophysiological study, Deephys' comparative analyses with artificial neural responses can enlighten the internal mechanisms about
% shine a flashlight into 
the black box of deep learning and generate insights into how or why a model has dysfunction in OOD cases.
\textit{Deephys} is an application, that is designed to provide a concrete guidance through its four main steps, to understand behavioral change of any deep model under distribution shift.
Even though we have experimented with object recognition models in this work, \textit{Deephys} can be applied to models specialized to other type of tasks i.e. object detection or segmentation.

An user should only provide the neural activations from any layer of a deep architecture, generated for different datasets according to the user's requirement, and the final layer logits (i.e. classification layer) to \textit{Deephys} for the analysis. These should also accompany images from all the selected datasets with their ground truth information as well as model predictions for complete visualizations.
The visualizations are spanned over multiple non-sequential steps consisting of individual neurons from the selected deep model, any image or a set of images, related to a specific category or a pair of categories confusing the model, from the datasets.
The users are expected to switch to the steps back and forth for complete understanding of the model functioning and it's inconsistent performance.
The steps of \textit{Deephys} are designed to provide easy navigation among them and efficient inspection of informative points for the users.

\textit{Deephys} provides concept-based explanations with neural responses by generating visualizations containing the top images from a dataset that maximally activate the neuron.
Such comparative visualizations can be efficiently generated through \textit{Deephys} with the flexibility to increase the number of top images for every neuron, that suits an user's choice.
Such visualization provides enriched insights about feature selectivity of the neurons and the model's perception of a dataset as a whole.
We also add possibility of explorations for every image through \textit{Deephys} in a meaningful way for the human observers. Please note that we consider only positive neural activations (by applying \textit{ReLu}) before transferring them to \textit{Deephys}.

Continuing the theme of comparative understanding between InD and OOD from the last step, we provide the similar `per \textbf{image}' visualization for OOD in \textit{Deephys}.  Please note that \textit{Deephys} allows an user to generate such comparative visualization including the activation normalization and activation ratio calculation, rather than interpreting with a single dataset, based on requirement towards the final goal.
% Please note that \textit{Deephys} allows the users to visualize more many top neurons for every image, but we select the top three neurons here to describe the idea.
\textit{Deephys} also allows the human observer to investigate a set of images for any category or any pair of categories following the explanation generation techniques shown in the last section. All these investigations are supported by quantitative metrics to measure feature novelty and spuriousness.

\subsection{Analyzing Colored MNIST}
With the purpose of investigating neural selectivity and their invariance to bias factors, we design a dataset by associating a color to every MNIST category. We are also interested to analyze the behavioral change of the neurons,
where we alter the association in meaningful ways with corresponding their effects in model performance.
Colors are added to each MNIST category such that all the images of a category are given a color and a different color is given to every category. We train a deep model with this dataset and hence consider this (Colored MNIST) as InD. We designed OOD versions by permuting the original colours to the categories (Permuted Colored MNIST), adding a new set of arbitrary colors (Arbitrary Colored MNIST), and adding a constant shift to all the original colors (Drifted Colored MNIST). Please refer Sec \ref{Method} for details on different versions of Colored MNIST.

Fig.\ref{fig:all_MNIST}(a) and (b) represent feature novelty in OOD versions compared to InD, that reveals neural tuning is mostly governed by the colors compared to shpae of the digits, and selectivity of the neuron change based on the colors. Adding constant shift to the colors show very small novelty as well as spuriousness, but arbitrary colors manifest high novelty and medium spuriousness. On the other hand, permuting the colors doesn't change the colors as a whole in OOD and neurons still fire for them associated with different categories, showing less novelty and high spuriousness. Fig.\ref{fig:all_MNIST}(b) and (d) show the similar traits for two sample neurons with top images from all the datasets. Neuron 21 fires for the same `dark yellow' color with less novelty (maximum activation ratio 83.6\%) compared to higher novelty (maximum activation ratio 23.4\%) for new color (from arbitrary set of colors). Neuron 39, on the other hand, is more tuned to `dark green' color and selective to different categories, both for Permuted and Arbitrary Colored MNIST (color similarity of `blue' to `green' and shape similarity of specific types of `0' and `5' in Fig.\ref{fig:all_MNIST}(d)), causing high spuriousness. Both the neurons show low novelty and spuriousness for Drifted Colored MNIST.

We presented two sample examples of mispredictions by the model, each from Permuted and Drifted Colored MNIST, with visualizations of the most activated neurons from the same dataset accompanied by visualizations of the same neurons from InD. This explains the behavior of the model about wrong predictions, as `0' from Permuted Colored MNIST is predicted as `2' due to neural tuning to `orange' color, but similarity of specific `2' shapes adds a little edge (57.3\% normalized activation of neuron 11) over similarity to specific `7' (52.1\% normalized activation of neuron 21). It is also noteworthy that the confusion of the model regarding shape is observed through neuron 14 which is the second most activated neuron (52.7\% normalized activation) and is tuned to specific shape related to `2' or `7'.

\subsection{Analyzing CIFAR10}
CIFAR10 contains natural images of 10 categories. Similar strategies were adopted to collect images for CIFAR10.2 consisting of the same categories. Please refer Sec \ref{Method} for details on these versions of CIFAR10. Even through the images are visibly similar, CIFAR10.2 achieves ~10\% less performance compared to CIFAR10, which is very hard to explain. Given that the main purpose of our proposed methodology is to explain natural datasets, we are more interested on the outcome when investigated with CIFAR10 versions. Fig\ref{fig:all_CIFAR}(a) and (c) clearly explains that there is little novelty as well as the neurons are selective to very similar categories for both the datasets. Considering specific neuron examples show similar trend, i.e. neurons respond to almost identical features such as `dog' (neuron 25) or `plane' (neuron 1), which are shown in Fig\ref{fig:all_CIFAR}(b) and (d). But analysing with the category confusions exhibits the important aspect of many mislabelled and difficult images in CIFAR10.2 compared to CIFAR10. Fig\ref{fig:all_CIFAR}(e) represents confusing images for 'plane' and 'dog' from both the datasets. Considering nearly similar scenarios for other pair of categories, CIFAR10.2 fails to reach similar performance compared to CIFAR10.

\subsection{Analyzing ImageNet}
Here, we select ResNet18 architecture and OOD versions of ImageNet (InD) with similar categories, but lesser number of images (ImageNetV2), sketchy images (ImageNet Sketch) or images
generated with added painting styles (Stylized ImageNet). Please refer Sec \ref{Method} for details on different versions of ImageNet.

We call different versions of ImageNetV2, ImageNet Sketch and stylized ImageNet as V2, sketch and style in short. While V2 contains natural images and is expected to exhibit similar behavior as InD, it has less novelty and spuriousness as well. Sketch is missing color features, but contains shape and texture features in most of the images, which can be misleading for the model and lead to misprediction. This is reflected in medium novelty and high spuriousness. On the other hand, style has additional painting information that leads to high novelty, but little less spuriousness. This analysis can be seen in Fig\ref{fig:all_IN}(a) and (c). We consider sample neurons to visually explain the plots in Fig\ref{fig:all_IN}(b) and (d). Neuron 33 shows high activation ratio for V2 and sketch representing high feature similarity or less novelty (110\% and 87.8\% respectively), but less similarity or high novelty for style (67\% ratio). Top images of neuron 177 show neural selectivity for very similar categories for V2 and little bit of style, but mostly different categories for sketch.

We show one example of misprediction by the model, from style, with visualizations of the most activated neurons from the same dataset accompanied by visualizations of the same neurons from InD in Fig\ref{fig:all_IN}(e). The most activated neuron for the selected `Bottlecap' image (neuron 243 with 65\% normalized activation) is tuned to images, that are mostly from `Honeycomb' category in InD. As the other neurons show comparatively lesser importance (i.e. 55.3\% and 54.8\% normalized activation for neuron 63 and 206 respectively) for the selected image, the model decision i.e. `Honeycomb' is mostly affected by neuron 243.

Very similar phenomenon for novelty and spurious scores is observed for ResNet50 which is presented in Fig\ref{fig:all_IN_R50}(a) and (b). We also experimented with the same network, but trained with style training data and considered as InD. Similar plots for this experimental setting are presented in Fig\ref{fig:all_IN_R50}(c) and (d), which show lower novelty and spuriousness for style as expected. Even though V2 has slightly more novelty than style, selectivity to different categories reflect more spuriousness. Sketch exhibits highest novelty and spuriousness that explains existence of very different features and association to different categories compared to style.

We also experiment with a convolutional vision transformer \cite{wu2021cvt} to understand if the neurons show any meaningful traits, similar to convolutional networks, and if analogous explanations as before can be generated for mispredictions in this case. 
% Here we consider output of the last convolutional transformer block (before the classification layer) and averaging through the spatial dimension is considered to provide activation scores for every neuron.
Fig\ref{fig:all_IN_CVT}(a) and (c) show very similar traits for novelty and spurious scores as ResNet18 and ResNet50 shown before.
Here, experimenting with the neuron visualization indicates that every neuron is tuned to more than one artifacts. Fig\ref{fig:all_IN_CVT}(b) visualizes sample neuron 31, that has highest novelty for style, but less novelty for V2 and sketch. Besides, the most activated images for neuron 6 in Fig\ref{fig:all_IN_CVT}(d) shows high selectivity to similar categories for V2, compared to sketch and style.
Fig\ref{fig:all_IN_CVT}(e) shows an `Electric locomotive' image, predicted by the model as `Passenger car', as the most activated neuron responds to `Passenger car' images from InD with high normalized activation (91.3\%) compared to subsequent neurons (67 and 233, with normalized activations 60.4\% and 57.3\% respectively).

\section{Discussion}

% \subsection{Explaining Distribution Shifts}
This ubiquitous phenomenon of nicely performing models failing under distribution shifts is well known to the community and has been an important research direction for many years now \cite{hendrycks2019benchmarking, szegedy2013intriguing, recht2019imagenet}. Multiple factors can trigger a distribution shift, such as a different data collection or preprocessing techniques, change in data source or environment \cite{koh2021wilds} and noise or small corruptions infused in the input data \cite{geirhos2018generalisation}. Prior works are mostly targeted for detecting distribution shifts \cite{rabanser2019failing, quinonero2008dataset, kulinski2020feature},
designing customized training techniques for improved generalization \cite{wang2020tent, sun2020test}, or applying data augmentation to the input data \cite{cubuk2018autoaugment, hendrycks2019benchmarking, hendrycks2019augmix, rusak2020simple, schneider2020improving, lopes2019improving}. Here, we focus on explaining distribution shifts through investigating the change in neural responses under such shifts with the purpose of making meaningful interpretations about inconsistent model behavior.

% \subsection{Neural Selectivity and Invariance Under Distribution Shifts}
Selectivity to a category or invariance to some features are two common facts, observed both for biological and artificial neurons, that are shown connected to OOD generalization \cite{anselmi2016invariance, sinha1996role, ullman2000high}.
While the factors of dataset bias are considered mostly a constrained set of orientations and illumination conditions \cite{alcorn2019strike}, deep networks are shown to overcome such biases by transferring the knowledge from richer set of conditions to a biased conditions for an object \cite{madan2022and, zaidi2020robustness}. A recent study showed that encouraging invariant neural representations can lead to improved OOD generalization for deep models \cite{sakai2022three}. Inability of all these approaches to be applied to natural datasets, due to absence of bias factors, instigated in suggesting a method to observe such biases by visualizing the change in individual neural responses.

% \subsection{Leveraging Visualization Tool}
Human intuition can play a critical role in identifying dataset biases that are otherwise not obviously quantifiable. It is possible to quantify after diligent attention given to them, but without the original intuition they may never have been found in the first place.
Despite being popular, but mainly a local explanation generation methods, feature attribution and gradient based approaches \cite{sundararajan2017axiomatic, ribeiro2016should} can be useful in interpreting failures from InD, where the same for OOD may not provide the most intuitive form of explanations for humans.
Also, these generated explanations may not efficiently capture visual disparities between InD and OOD inputs \cite{adebayo2020debugging}. 
For this work, we consider concept-based explanations to interpret model behaviours in terms of concepts or artifacts that are more aligned to human understanding. Recent concept-based explainability techniques mainly focus on tweaking the object recognition based models to generate better explanation \cite{koh2020concept, sarkar2022framework}, or explain the decision of OOD detectors \cite{choi2022concept}.

Some of the previous efforts tried to improve model generalization capability by minimizing the dependency on spurious features or improving the effect of other non-spurious features.
\cite{arjovsky2019invariant} proposed a new model training approach that reduces the statistical association to spurious features by encouraging model invariance to them.
\cite{eastwood2021source} showed that restoring important features from InD and extracting features with the same semantics from OOD can improve model performance against OOD.
Our approach, on the other hand, follows concept-based visualizations to identify the features learnt by individual neurons, both for InD and OOD datasets.
We leverage such comparative visualizations to investigate spuriousness of the features in neuron level and elucidate the cause of abnormal model behaviors.
% On the other hand, our approach is applicable for interpreting the behavioural change of any deep network under distribution shifts, where we follow concept-based explanations based on the network's neural responses for InD and OOD datasets.
% We leverage such explanations to generate synchronized visualizations that assists in investigating biases for all the datasets and elucidating the cause of abnormal model behaviors.

\begin{figure*}[h]
    \centering
    \begin{tabular}{c}
    \includegraphics[width=\textwidth]{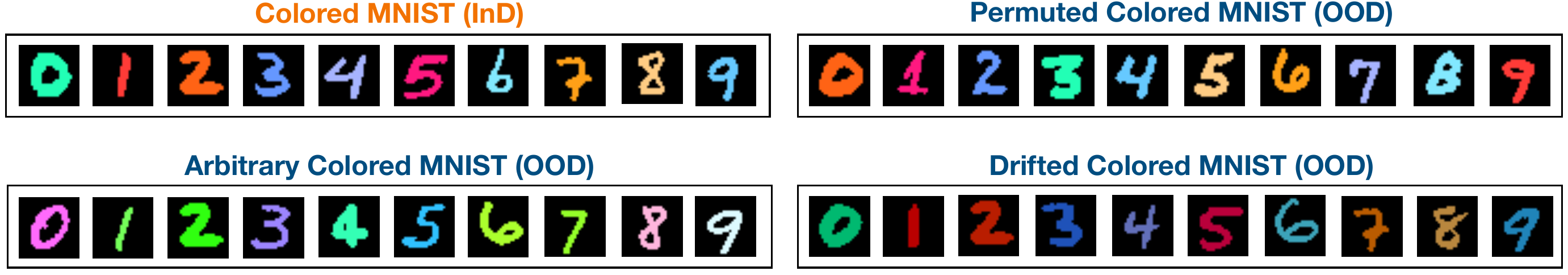} \\
    (a) \\ [1em]
    \includegraphics[width=\textwidth]{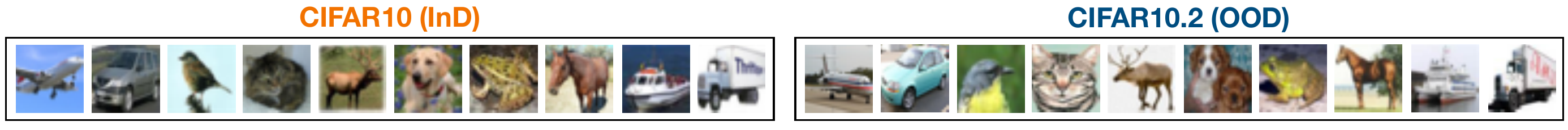} \\
    (b) \\ [1em]
    \includegraphics[width=\textwidth]{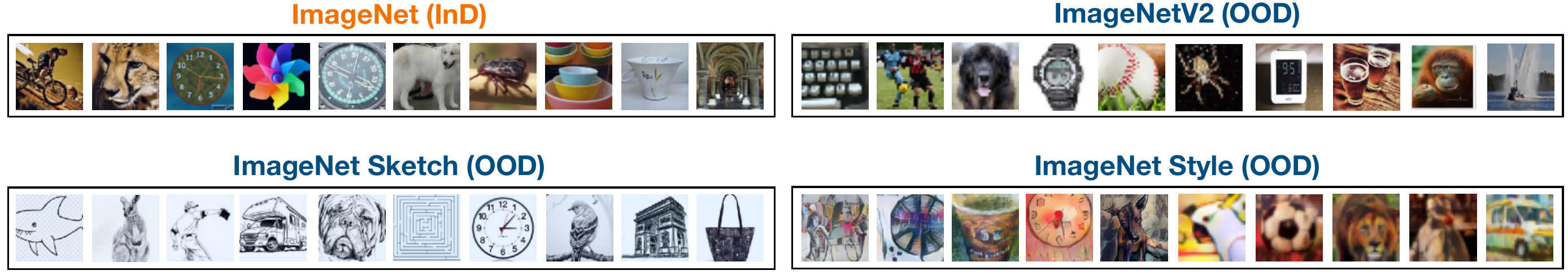} \\
    (c) \\
    \end{tabular}
    \caption[XYZ]{\label{fig:all_datasets} Examples of all the InD and OOD datasets considered for experiments. (a) Different colors are added to each MNIST category and then used to train a model. Other OOD versions are generated by permuting the colours (Permuted Colored MNIST), adding arbitrary colors (Arbitrary Colored MNIST), and adding a constant shift to all the original colors (Drifted Colored MNIST). (b) CIFAR10 contains natural images of 10 categories. Similar strategies were adopted to collect images for CIFAR10.2 consisting of the same categories. (c) ImageNet is comparatively much larger and more complex dataset with natural images from 1000 categories. All the OOD versions of ImageNet have the same categories, but contain lesser number of images (ImageNetV2), sketch images (ImageNet Sketch) or images generated with added painting styles (Stylized ImageNet).}
\end{figure*}

\section{Methods}
\label{Method}

\subsection{Datasets}

A very specific set of datasets are considered for the experiments to exhibit the effectiveness of Deephys. Even though we claim that Deephys can be applied for datasets with natural images, we are also interested in datasets with manually added bias factors. We present the datasets below and show how Deephys can generate insights about failure cases of a model.

\subsubsection{Colored MNIST}

MNIST is a handwritten digit dataset comprising of single channel images from 10 categories, 0 to 9. Different colors are added to each MNIST category such that all the images of a category are given a color and a different color is given to every category. We train a deep model with this dataset and hence consider this (Colored MNIST) as InD.
Similar idea is followed for the OOD versions, but they differ in the set of colors that are selected for each category.
We generate three versions by permuting the original colours to the categories (Permuted Colored MNIST), adding a new set of arbitrary colors (Arbitrary Colored MNIST), and adding a constant shift to all the original colors (Drifted Colored MNIST). Please refer to part (a) of Fig.\ref{fig:all_datasets} for example images from all the InD and OOD versions of Colored MNIST.

\subsubsection{CIFAR-10 Versions}

CIFAR10 contains natural images of 10 categories. Similar strategies were adopted to collect images for CIFAR10.2 consisting of the same categories. Sample images from all the categories of both the datasets are shown in part (b) of Fig.\ref{fig:all_datasets}.

\subsubsection{ImageNet Versions}

ImageNet is larger and more complex dataset, compared to CIFAR10, containing natural images from 1000 categories. All the OOD versions of ImageNet have the same categories, but contain lesser number of images (ImageNetV2), sketchy images (ImageNet Sketch) or images
generated with added painting styles (Stylized ImageNet). Sample images from all ImageNet and the OOD versions are presented in part (c) of Fig.\ref{fig:all_datasets}.

\subsection{Network Architectures}
We considered a small 3 layer convolutional network for experiments with Colored MNIST and studied the penultimate layer with 50 neurons for our analyses. For CIFAR10, we considered a ResNet18 architecture, that has 512 neurons in the penultimate layer. With the purpose of generating more human understandable explanations with a dataset containing only 10 categories, we added a fully connected layer, with 50 neurons, between the penultimate layer and the classification layer of the ResNet18 architecture.

ImageNet pretrained ResNet18 and ResNet50 architectures are considered for all our experiments related to ImageNet and it's OOD versions. We also studied model behaviour with a convolutional vision transformer \cite{wu2021cvt} to check if similar traits are observed here. We consider output of the last convolutional transformer block (before the classification layer) and averaging through the spatial dimension is considered to provide activation scores for every neuron.

\subsection{Analysing the OOD Failures through Neural Activity Comparison}
The different visualization-based toolsets of deephys, which are a way to realize the behavioral disparities of the model between InD and OOD, are mainly based on the neural activations of a model captured with all the datasets. 
Considering that neurons of a deep model responds to specific artifacts from InD, their responses change based on similarity of the features in OOD.
We are not able to quantity such similarity of features across datasets, but can we make estimations based on neural responses for them.
Here, we analyze the activations through different quantitative measures, which are governed by our hypothesis that there are mainly two possible scenarios where the model neurons can manifest different behaviours.
These scenarios can trigger a change in selectivity or invariance of the neurons and, in turn, degrade model performance under distribution shift.
Such quantitative analyses provides a global picture of neural response comparison between InD and OOD, that can complement the understanding about these datasets acquired through deephys.
% It is worth mentioning that investigating any layer of a model can provide important insights about it's behavior, and can be performed through deephys. But, we only consider the penultimate layer for this study due to the closest proximity to the decision layer, that may reveal the most differentiating traits at the neuron level.
% Hence, comparing the neural responses for InD and OoD provides new insights about the type of distribution shift that can complement the understanding acquired through deephys.

\subsection{Novel features in OOD}
As feature detectors, the neurons respond to patterns originating from InD. They respond When they encounter similar patterns in OOD, but it depends on the quantity of the feature similarity which can be measured by the difference in the firing rates for InD and OOD. We present a metric, '\textbf{novelty score}' that can indicate the originality found by neurons in OOD compared to InD.
We calculate the average category-wise activations for every neuron and select the highest average scores i.e. the highest average activations of the neurons for any category.
Please note that the category is not important here, rather the highest response for any category is studied.
The score for OOD is then subtracted from the score for InD for every neuron, that provides novelty score for them.
We only consider the neurons whose subtracted score is positive i.e. that signifies some novel information found by a neuron in OOD, which is not present in InD.
A density plot is drawn with these final values and presented as novelty score for OOD with respect to InD.

% \subsubsection{Novelty Score}

\subsection{Spurious features in InD}
Neurons, that are tuned to a spurious feature, can still be activated by the same feature in the OOD. When the spurious feature is associated to a different categories in the OOD, the neuron shows high response to the spurious feature but also become selective to the other category. We devise a metric, `\textbf{spurious score}' to capture the said effect.
Here we apply spearman correlation between the average category-wise activations, calculated in the same way as for the `novelty score', of OOD and InD. Absolute values of the correlation scores are subtracted from unity and plotted in a density plot to obtain `spurious score'. The purpose of such calculation is that a lesser score would represent less spuriousness i.e. nerons fire for similar categories with OOD as InD, and related explanation can be presented for high score. 

% Once we have categorywise average firing rates of every neuron, both for InD and OOD, we calculate spearman correlation between them and subtract the absolute correlation scores from unity to come up with spurious score. This reflects the amount of spuriousness an OOD dataset has compared to InD i.e. how much neurons respond to different categories on an average for the considered datasets. 

% \section{Conclusion and Future Work}
% We propose a method to assist researchers understand how changes in the distribution of the input data can affect model performance and decision-making. We design a software that serves the said purpose for any object recognition network with the in-distribution data and the data from other distribution, where the model shows inconsistent behavior. We also show effectiveness of the software ,with some well-known datasets and their OOD versions, and different network architectures. We hope the proposed software and the ideas will be helpful to unveil abnormal model behavior under different aspects including adversarial robustness and corruptions.

\section*{Author contributions}
AS and XB designed research with contributions of IM; AS performed experiments with contributions of XB; MG developed the app; MG and XB prepared the documentation and the website, with contributions of AS; AS wrote the paper with contributions of all the rest; XB and TS supervised the research. 

\section*{Data availability}
The code to generate Colored MNIST is contained in 'tutorials' sub-folder under the following GitHub repository: \url{https://github.com/mjgroth/deephys-aio}. All the other datasets (both InD and OOD versions) are publicly available.

\section*{Code availability}
The website of the project is \url{https://deephys.org}.
The source code of Deephys Application is publicly available at the same GitHub repository.
The code to reproduce the results, reported in this paper, is contained in 'tutorials' sub-folder under the repository.
% Spefic link to the code to reproduce results ('tutorials' sub-folder under full repo).

\section*{Acknowledgements}
We are grateful to Tomaso Poggio, Pawan Sinha, Serban Georgescu and Hisanao Akima for their insightful advice and warm encouragement. This work was supported by 
Fujitsu Limited (Contract No.~40009568).

\section*{Conflicts of Interests Statement}
The authors declare that the research was conducted in the absence of any commercial or financial relationships that could be construed as a potential conflict of interest. Fujitsu Limited funded this study  (Contract No.~40009568) and also participated in the study through TS. All authors declare no other competing interests.

% \begin{center}

\bibliography{example_paper}
\bibliographystyle{icml2022}

%%%%%%%%%%%%%%%%%%%%%%%%%%%%%%%%%%%%%%%%%%%%%%%%%%%%%%%%%%%%%%%%%%%%%%%%%%%%%%%
%%%%%%%%%%%%%%%%%%%%%%%%%%%%%%%%%%%%%%%%%%%%%%%%%%%%%%%%%%%%%%%%%%%%%%%%%%%%%%%
% APPENDIX
%%%%%%%%%%%%%%%%%%%%%%%%%%%%%%%%%%%%%%%%%%%%%%%%%%%%%%%%%%%%%%%%%%%%%%%%%%%%%%%
%%%%%%%%%%%%%%%%%%%%%%%%%%%%%%%%%%%%%%%%%%%%%%%%%%%%%%%%%%%%%%%%%%%%%%%%%%%%%%%
\newpage
\appendix
\onecolumn

\end{document}